\documentclass{article}

\PassOptionsToPackage{numbers, compress}{natbib}
\usepackage[final]{neurips_2024}

\usepackage[utf8]{inputenc} 
\usepackage[T1]{fontenc}    
\usepackage{hyperref}       
\usepackage{url}            
\usepackage{booktabs}       
\usepackage{amsfonts}       
\usepackage{nicefrac}       
\usepackage{microtype}      
\usepackage{xcolor}         

\usepackage{graphicx}
\usepackage{amsmath}
\usepackage{amssymb}
\usepackage{booktabs}
\usepackage{multirow}
\usepackage{threeparttable}
\usepackage{bbding}
\usepackage{makecell}
\usepackage{subcaption}
\usepackage{xspace}
\usepackage{xcolor}
\usepackage{colortbl}
\usepackage{tcolorbox}
\usepackage{wrapfig}
\usepackage[misc]{ifsym}

\definecolor{mycolor_blue}{HTML}{E7EFFA}
\definecolor{mycolor_green}{HTML}{E6F8E0}
\definecolor{mycolor_red}{HTML}{F9C1BD}
\definecolor{mycolor_orange}{HTML}{F4D9C4}
\definecolor{mycolor_gray}{HTML}{D9D9D9}
\definecolor{mycolor_purple}{HTML}{EDD5FD}
\definecolor{mycolor_yellow}{HTML}{FCEEBE}
\definecolor{dt}{gray}{0.6}
\newcommand{\VarSty}[1]{\textnormal{\ttfamily\color{black}#1}\unskip}
\newcommand{\authorskip}{\hspace{4mm}}

\title{MoVA: Adapting Mixture of Vision Experts to Multimodal Context} 

\author{Zhuofan Zong$^{1,2,}$\thanks{Equal contribution.} \authorskip  Bingqi Ma$^{2,*}$ \authorskip Dazhong Shen$^3$ \authorskip Guanglu Song$^2$ \\[2ex] \textbf{Hao Shao}$^1$ \authorskip \textbf{Dongzhi Jiang}$^1$ \authorskip \textbf{Hongsheng Li}$^{1,3,4,}$\thanks{Corresponding authors.} \authorskip \textbf{Yu Liu}$^{2,\dag}$ \\\\
$^1$CUHK MMLab \authorskip $^2$SenseTime Research \authorskip $^3$Shanghai AI Laboratory \authorskip $^4$CPII under InnoHK
}

\begin{document}
\maketitle
\begin{abstract}
As the key component in multimodal large language models~(MLLMs), the ability of the visual encoder greatly affects MLLM's understanding on diverse image content.
Although some large-scale pretrained vision encoders such as vision encoders in CLIP and DINOv2 have brought promising performance, we found that there is still no single vision encoder that can dominate various image content understanding, e.g., the CLIP vision encoder leads to outstanding results on general image understanding but poor performance on document or chart content.
To alleviate the bias of CLIP vision encoder, we first delve into the inherent behavior of different pre-trained vision encoders and then propose the MoVA, a powerful and novel MLLM, adaptively routing and fusing task-specific vision experts with a coarse-to-fine mechanism.
In the coarse-grained stage, we design a context-aware expert routing strategy to dynamically select the most suitable vision experts according to the user instruction, input image, and expertise of vision experts.
This benefits from the powerful model function understanding ability of the large language model~(LLM).
In the fine-grained stage, we elaborately conduct the mixture-of-vision-expert adapter~(MoV-Adapter) to extract and fuse task-specific knowledge from various experts.
This coarse-to-fine paradigm effectively leverages representations from experts based on multimodal context and model expertise, further enhancing the generalization ability.
We conduct extensive experiments to evaluate the effectiveness of the proposed approach.
Without any bells and whistles, MoVA can achieve significant performance gains over current state-of-the-art methods in a wide range of challenging multimodal benchmarks.
Codes and models are available at \url{https://github.com/TempleX98/MoVA}.

\end{abstract}
\section{Introduction}
\label{sec:intro}

Significant achievements in multimodal large language models (MLLMs)~\cite{blip2, flamingo, llava, internvl, shikra, qwenvl, instructblip} have been witnessed due to their remarkable proficiency in solving open-world tasks.
MLLMs acquire visual perception capacity while inheriting sophisticated reasoning abilities and knowledge from large language models (LLMs)~\cite{vicuna,qwen,zhang2024map}.
The core idea behind MLLMs is projecting the vision representation into an LLM through a projector, facilitating a general-purpose multimodal understanding.

General multimodal understanding requires comprehending complex image contexts across various tasks and scenarios.
The CLIP~\cite{clip} vision encoder, pre-trained on large-scale image-text pairs with a contrastive loss, is widely considered as a flexible and popular choice among the latest leading MLLMs.
However, training data and optimization target of the vision encoder determine its inconsistent performance across tasks and scenarios, which will bias the generalization of multimodal large language models.
For instance, MLLMs with a single CLIP vision encoder usually perform poorly on fine-grained tasks such as grounding and optical character recognition~(OCR)~\cite{vary}.
Several works have attempted to incorporate extra state-of-the-art vision encoder experts to cope with the challenge.
For example, both SPHINX~\cite{sphinx} and MoF~\cite{mof} integrate vision self-supervised learning features of DINOv2~\cite{dinov2} with MLLMs to enhance their visual grounding capabilities.
Vary~\cite{vary} introduces a new vision encoder expert for improved fine-grained document and chart parsing ability.
Intuitively, it is necessary to explore the utilization of more task-specific vision encoder experts in MLLMs to promote model generalization across various domains.

\begin{table}[t]
    \centering
    \caption{\textbf{Comparison of CLIP \textit{vs.} state-of-the-art task-specific vision encoders.} Our evaluation criteria encompass a variety of dimensions: comprehensive benchmarks~\cite{mmbench}, text-oriented Visual Question Answering (VQA)~\cite{docvqa,chartqa}, general VQA~\cite{gqa}, object hallucination~\cite{pope}, Referring Expression Comprehension (REC)~\cite{refcoco}, Referring Expression Segmentation (RES)~\cite{refcoco}, and medical VQA benchmark SLAKE~\cite{slake}. We use the same data for each model.}

  \label{tab:intro_exp}
  \setlength\tabcolsep{2pt}
  \resizebox{\textwidth}{!}{
  \begin{tabular}{@{}c|c|cccccccc@{}}
    \toprule
    Vision Encoder & Task & MMB & \cellcolor{mycolor_orange}DocVQA & \cellcolor{mycolor_gray}ChartQA & \cellcolor{mycolor_blue}GQA & \cellcolor{mycolor_green}POPE & \cellcolor{mycolor_blue}REC & \cellcolor{mycolor_red}RES & \cellcolor{mycolor_purple}SLAKE \\
    \midrule
    CLIP~\cite{clip} & Image-text Contrastive & \bf{64.9} & 35.6 & 35.3 & 62.5 & 85.7 & 81.5 & 43.3 & 63.7 \\
    \midrule
    \cellcolor{mycolor_blue}DINOv2~\cite{dinov2} & \cellcolor{mycolor_blue}Visual Grounding & 57.5 & 14.7 & 15.9 & \cellcolor{mycolor_blue}\bf{63.9} & 86.7 & \cellcolor{mycolor_blue}\bf{86.1} & 47.5 & 59.4 \\
    \cellcolor{mycolor_green}Co-DETR~\cite{codetr} & \cellcolor{mycolor_green}Object Detection & 48.4 & 14.2 & 14.8 & 58.6 & \cellcolor{mycolor_green}\bf{88.0} & 82.1 & 48.6 & 55.3 \\
    \cellcolor{mycolor_red}SAM~\cite{sam} & \cellcolor{mycolor_red}Image Segmentation & 40.7 & 13.9 & 15.0 & 54.0 & 82.0 & 79.2 & \cellcolor{mycolor_red}\bf{49.3} & 57.7 \\
    \cellcolor{mycolor_orange}Pix2Struct~\cite{pix2struct} & \cellcolor{mycolor_orange}Text Recognition & 41.9 & \cellcolor{mycolor_orange}\bf{57.3} & \cellcolor{mycolor_orange}53.4 & 51.0 & 78.1 & 59.2 & 32.2 & 44.0 \\
    \cellcolor{mycolor_gray}Deplot~\cite{deplot} & \cellcolor{mycolor_gray}Chart Understanding & 36.2 & 40.2 & \cellcolor{mycolor_gray}\bf{55.8} & 48.1 & 75.6 & 51.1 & 27.0 & 44.5 \\
    \cellcolor{mycolor_yellow}Vary~\cite{vary} & \cellcolor{mycolor_yellow}Document Chart Parsing & 28.1 & \cellcolor{mycolor_yellow}47.8 & \cellcolor{mycolor_yellow}41.8 & 42.6 & 69.1 & 21.6 & 16.0 & 40.9 \\
    \cellcolor{mycolor_purple}BiomedCLIP~\cite{biomedclip} & \cellcolor{mycolor_purple}Biomedical Contrastive  & 40.0 & 15.3 & 16.8 & 50.8 & 76.9 & 57.8 & 27.4 & \cellcolor{mycolor_purple}\bf{65.1} \\
    \midrule
    Plain fusion & - & 63.4 & 46.5 & 48.9 & 63.0 & 86.4 & 85.7 & 45.3 & 64.7 \\
    \cellcolor{gray!15}MoVA & \cellcolor{gray!15}- & \cellcolor{gray!15}\bf{65.9} & \cellcolor{gray!15}\bf{59.0} & \cellcolor{gray!15}\bf{56.8} & \cellcolor{gray!15}\bf{64.1} & \cellcolor{gray!15}\bf{88.5} & \cellcolor{gray!15}\bf{86.4} & \cellcolor{gray!15}\bf{49.8} & \cellcolor{gray!15}\bf{66.3} \\    
  \bottomrule
  \end{tabular}
  }
\end{table}

We aim to start the exploration through empirical analysis of readily available vision experts. 
In particular, we focus on the multimodal capabilities of seven distinct state-of-the-art vision encoders based on LLaVA-1.5-7B~\cite{llava15}.
The results in Table~\ref{tab:intro_exp} reveal that MLLMs with these task-specific vision encoders achieve optimal performance in their respective area.
Concurrently, we note that the plain fusion~(concatenation) of vision encoder experts adopted in previous works~\cite{sphinx} would not bring consistent improvement compared with the single task-specific vision expert in its proficient task.
The inherent bias of each expert introduces biased information and leads to performance degradation in the plain fusion paradigm. 
For example, DINOv2 serves as an expert in visual grounding but performs poorly at text-oriented tasks.
Representation of DINOv2 would be regarded as biased information in text-related scenarios so incorporating DINOv2 for these tasks would inevitably cause performance decrease.
Consequently, a flexible method of vision encoder ensemble that dynamically activates and weights context-relevant task-specific vision experts can fully unleash the capacity of these models while avoiding model bias.

In this paper, we propose MoVA, a powerful MLLM, adaptively routing and fusing task-specific vision experts with a coarse-to-fine mechanism.
Inspired by the powerful tool-use capabilities of LLM~\cite{toolllm}, the coarse-grained context-aware expert routing aims to employ LLM to select vision experts with strong relevance to the user's image and instruction from the expert model pool.
Thanks to the strong generalization ability of LLM, we also can perform model routing for vision experts in open scenarios.
The fine-grained expert fusion facilitates better extraction and integration of expert representations based on multimodal context.
Specifically, the expert knowledge extractor in the mixture-of-vision-expert adapter (MoV-Adapter) will extract diverse task-specific knowledge from various vision experts through mixture-of-expert (MoE) cross-attention layers.
The dynamic gating network can allocate precise expert-wise soft weights for the integration of extracted task-specific knowledge.
Under the coarse-to-fine paradigm, we provide a flexible and effective manner of leveraging representation from experts based on multimodal context and model expertise, further enhancing the model generalization ability.
As presented in Table~\ref{tab:intro_exp}, MoVA can preserve the optimal performance of a single relevant vision encoder by ignoring non-relevant experts on the GQA, POPE, and REC task.
Besides, MoVA can further boost performances via the fine-grained fusion of multiple relevant vision experts on other tasks.

We conduct comprehensive experiments on various benchmarks to evaluate the effectiveness of MoVA, including MLLM benchmarks, visual question answering~(VQA), visual grounding, and biomedical understanding.
Without any bells and whistles, MoVA can achieve significant performance gains over current state-of-the-art methods.

The \textbf{contributions} of this work are three-fold: \textbf{(i)} By analyzing the performance of individual vision encoders versus the plain fusion of multiple encoders across various tasks, we reveal that the inherent bias of each vision encoder can diminish its generalization ability across other irrelevant domains. \textbf{(ii)} We propose MoVA, a powerful MLLM composed of coarse-grained context-aware expert routing and fine-grained expert fusion with MoV-Adapter.
Based on multimodal context and model expertise, MoVA fully leverages representation from multiple context-relevant vision experts flexibly while avoiding biased information of irrelevant experts. \textbf{(iii)} We demonstrate the effectiveness of each component in MoVA by elaborate ablation studies. 
MoVA can achieve significant performance gains over state-of-the-art methods in a wide range of challenging benchmarks.
\section{Related Work}

Multimodal architectures~\cite{blip2, llava, qwenvl, lmdrive, visualcot, jiao2024lumen, ma2024exploringrolelargelanguage,guo2023owl}, optimization paradigm~\cite{zhou2024aligning,zhou2024calibrated}, applications~\cite{jiang2024comat,xia2024rule,xia2024mmed,yang2024boosting}, and benchmarks~\cite{yue2024mmmu,li2024eyes,zhang2024eventhallusion,mmsearch,xia2024cares,xia2024mmie,wang2023rolellm} have recently achieved remarkable progress and garnered unprecedented attention within the academic community.
Multimodal large language models (MLLMs) usually leverage the alignment from visual features to the linguistic feature space to achieve superior vision-language understanding capabilities based on off-the-shelf LLMs and vision encoders.
CLIP vision encoder~\cite{clip}, which is trained in contrastive learning from billions of diverse image-text pairs~\cite{laion400m, laion5b}, is widely used among these works.
For example,
LLaVA~\cite{llava} adopts an MLP projector to align visual tokens from the frozen CLIP vision encoder to the embedding layer of LLM.
However, The representation from CLIP exhibits strong discriminative abilities in classification and recognition but only has limited performance on downstream tasks like location and relation understanding~\cite{t2ibench}.
To break through this bottleneck, some works~\cite{internvl,sharegpt4v} turn to unlock the CLIP vision encoder and further fine-tune the parameter with training data for downstream tasks.
For instance,
Qwen-VL~\cite{qwenvl} collected massive training data for grounding and OCR to jointly optimize the CLIP vision encoder and LLM.
Recent works propose to involve an extra frozen vision encoder to enhance the performance of MLLMs.
SPHINX~\cite{sphinx} is one of the pioneers, where grounding capabilities have been significantly improved with the assistance of the DINOv2~\cite{dinov2}.
Vary~\cite{vary} introduces an extra encoder training on large-scale charts and document data to improve the performance on related downstream tasks.

\section{MoVA Methodology}
\label{sec:method}
\begin{figure*}[tp]
    \centering
    \includegraphics[width=0.99\textwidth]{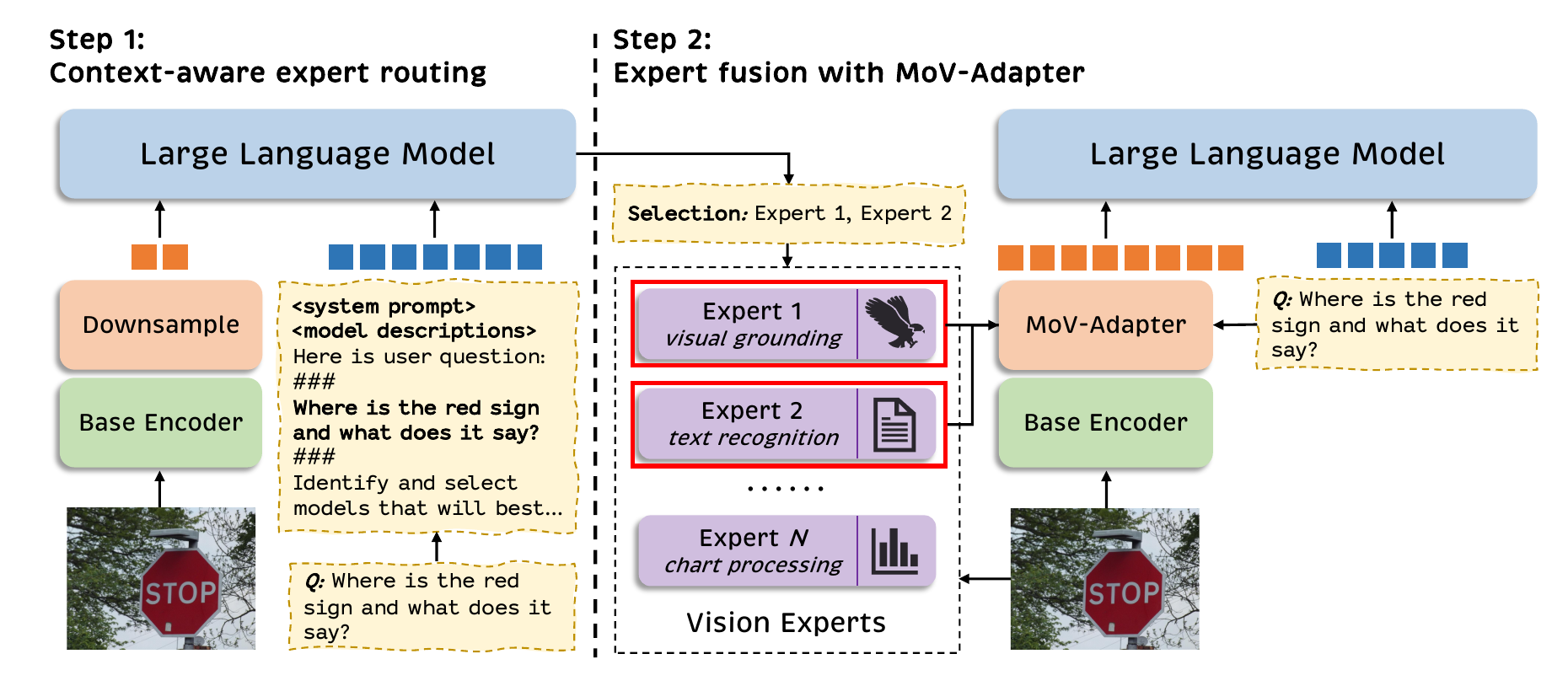}
    \caption{\textbf{The pipeline of MoVA.} MoVA performs coarse-to-fine routing to solve a given question. The coarse context-aware expert routing is performed in the first stage to select context-relevant experts. Next, we adopt the MoV-Adapter to extract and fuse the task-specific knowledge from these selected experts in a fine-grained manner.}
    \label{fig:framework}
\end{figure*}

\subsection{Overview}
MoVA comprises five key components: \textbf{(i)} a pre-trained large language model (LLM) that generates accurate responses given the image tokens and instructions; \textbf{(ii)} a base vision encoder; \textbf{(iii)} vision experts that generate task-specific vision latent features; 
\textbf{(iv)} mixture-of-vision-expert adapter (MoV-Adapter) that performs fine-grained expert fusion based on the multimodal context.

As illustrated in Figure~\ref{fig:framework}, MoVA consists of two stages: coarse-grained context-ware expert routing and fine-grained expert fusion with MoV-Adapter.
First, our coarse-grained context-ware expert routing leverages the tool-use capabilities of LLM, routing the most appropriate experts from $N$ expert candidates via LLM to help the model answer the user's question.
In the second stage, we turn to enhance the visual representation with a novel MoV-Adapter module in a fine-grained manner.
More specifically, we leverage the mixture-of-expert (MoE) cross-attention layers to extract the task-specific knowledge of representations from chosen experts.
Meanwhile, the dynamic gating network in MoV-Adapter can allocate soft weights to the extracted knowledge of each expert according to the input image and instruction.
Then the extracted knowledge can be effectively integrated into the foundational representation of the base vision encoder.
Finally, the enhanced visual representation with instruction tokens is fed to the LLM to generate an accurate response.
In Section~\ref{sec:context_routing} and Section~\ref{sec:mov_adapter}, we will focus on our core contributions, the context-aware expert routing strategy, and the expert fusion with MoV-Adapter.
In Section~\ref{sec:training}, we will introduce the training process.

\textbf{Pretrained Vision Encoders and LLM.}
The vision encoders in MoVA consist of a base encoder and multiple task-specific vision encoder experts.
We choose the pre-trained CLIP ViT-L-336px as the base encoder.
Our vision experts include several state-of-the-art task-specific encoders: DINOv2, Co-DETR, SAM, Pix2Struct, Deplot, Vary, and BiomedCLIP.
The corresponding expertise is presented in Table~\ref{tab:intro_exp}.
For example, both Pix2Struct and Vary will be used when the user asks the MLLM to scan the document image.
MoVA is flexible and easy to generalize to all decoder-only LLMs.
We mainly consider Vicuna-7B~\cite{vicuna}, Llama3-8B~\footnote{https://github.com/meta-llama/llama3}, and Yi-34B~\cite{young2024yi} as our language models in this work.

\subsection{Coarse-grained Context-aware Expert Routing} 
\label{sec:context_routing}

\textbf{Pipeline of Context-aware Routing.}
The context-aware expert routing strategy aims to employ the impressive tool-use capacity of LLM to select vision experts with strong relevance to the user's image and instruction from a model pool.
Specifically, we perform the context-aware expert routing in three steps during inference.
First, the input image, user questions, and descriptions of expert models are converted into appropriate instructions that prompt the MLLM to perform expert selection.
An example of the prompt instruction input and selection output is shown in Table~\ref{tab:example_prompt}.
Such a routing task does not require image details and high-resolution input images, hence we directly downsample the base encoder's visual feature to obtain a coarse image embedding (\textit{e.g.}, $144$ image tokens).
The downsampled image tokens and instruction tokens are then fed to the LLM as inputs.
Finally, the LLM generates the output text and we parse it to determine which vision expert should be selected for fine-grained knowledge extraction in the second stage.
For instance, as depicted in Table~\ref{tab:example_prompt}, the LLM directly outputs the option's letter of DINOv2 and Pix2Struct, thus we only utilize them for the subsequent extraction.
During training, we do not perform context-aware expert routing and replace the routing outputs with our routing annotations to improve efficiency.

\begin{table*}[t!]
\centering
\caption{One example of the instruction-following data for context-aware expert routing. We present the multimodal inputs in the top block and the language response in the bottom block. The detailed model descriptions are released in the Appendix.}
\setlength\tabcolsep{3pt}
\begin{minipage}{0.99\columnwidth}
\vspace{0mm}    
\centering
\begin{tcolorbox} 
    \centering
      \footnotesize
    \begin{tabular}{p{0.97\columnwidth} c}
   \VarSty{ {\bf {\underline{Routing Prompt Input}}} } &\\
You are a helpful assistant router. Based on the visual content, questions, and model pool the user provides, you need to consider the expertise of these models to select the most 3 suitable models to help you answer the questions. Answer with the model's letter from the given choices directly. If no models are selected, just answer 'none'.
 &  \\
Model pool:&  \\
A. <DINOv2 model description> & \\
B. <Co-DETR model description> & \hspace{-3.2cm} \multirow{5}{*}{ \includegraphics[height=2.0cm]{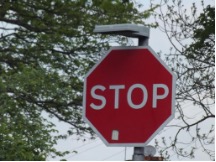} } \\
C. <SAM model description> & \\
D. <Pix2Struct model description> & \\
E. <Deplot model description> & \\
F. <Vary model description> & \\
G. <BiomedCLIP model description> & \\
Question: & \\ 
Where is the red sign and what does it say? & \\ 
    \hrulefill & \\
   \VarSty{ {\bf \underline{Routing Prompt Output}} } & \\
A, D& \\
    \end{tabular}
\end{tcolorbox}
    \label{tab:example_prompt}    
\end{minipage}
\end{table*}

\textbf{Routing Data Construction.}
\label{sec:routing_data}
Compared with other MLLMs, MoVA requires additional routing annotations.
We first introduce the formal definition of the data structure for an unambiguous understanding of the routing data.
The data structure for expert routing introduces additional routing annotation $\mathcal{R}$ to the conventional multimodal data $(\mathcal{I},\mathcal{Q},\mathcal{A})$.
Here, $\mathcal{I}$ represents the image, $\mathcal{Q}$ and $\mathcal{A}$ refer to the question-answer pair, and $\mathcal{R}$ refers to the expert set which contains the most appropriate ones to solve this question.
Then the construction process for routing data can be formulated as $(\mathcal{I},\mathcal{Q},\mathcal{A}) \rightarrow \mathcal{R}$, with the primary objective being to derive vision experts that optimally align with the sample $(\mathcal{I},\mathcal{Q},\mathcal{A})$.
Intuitively, the language modeling loss can serve as an effective metric for evaluating how a data sample aligns with the vision expert.
Specifically, we can reuse the LLaVA-1.5-7B models with various vision encoders presented in Section~\ref{sec:intro} to perform loss computation.
Here, we denote the model with the base encoder as $\mathcal{M}_0$  and the model with $j$-th expert among $N$ experts as $\mathcal{M}_j$.
For the $i$-th sample $(\mathcal{I}_{i},\mathcal{Q}_{i},\mathcal{A}_{i})$, we send it to models~$\{\mathcal{M}_j|j \in \{0, 1, \ldots, N\}\}$ and calculate the language modeling loss~$\{\mathcal{L}_{i,j}|j \in \{0, 1, \ldots, N\}\}$.
The $j$-th expert is regarded as a useful expert for the $i$-th sample only if $\mathcal{L}_{i, j} < \mathcal{L}_{i, 0}$ and will be added to the routing set $\mathcal{R}_{i}$.
Note that we only keep up to 3 vision experts to avoid computation costs brought by too many additional experts.
All the routing annotations of our training data are generated offline.
We can directly parse and input these offline results to the subsequent expert fusion component during training.

\textbf{Routing Data Augmentation.}
To preserve the expert routing robustness and generalization ability in open scenarios, we only randomly select 2K samples for training, remove the model name in model description, and rewrite the model descriptions using ChatGPT~\cite{gpt4} for each expert.
We also shuffle the model pool and randomly truncate the model pool during training. 

\subsection{Fine-grained Expert Fusion with MoV-Adapter}
\label{sec:mov_adapter}
\begin{wrapfigure}[13]{r}{0.36\linewidth}
    \centering
    \includegraphics[width=\linewidth]{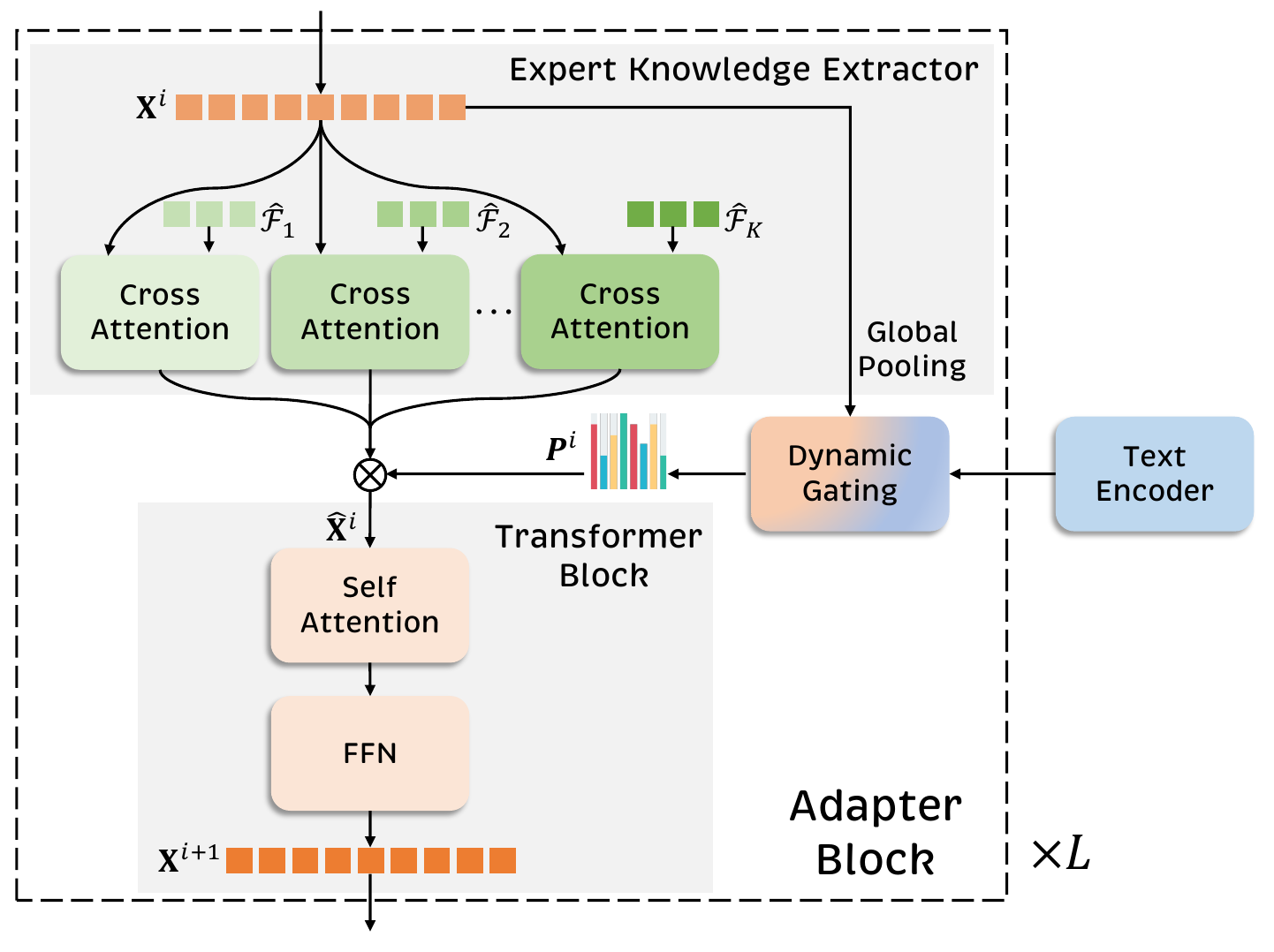}
    \caption{MoV-Adapter architecture.
    }
    \label{fig:adapter}
\end{wrapfigure}

We propose the MoV-Adapter to facilitate fine-grained expert representation extraction and integration based on multimodal context.
As shown in Figure~\ref{fig:adapter}, the MoV-Adapter consists of~$L$ adapter blocks and a text encoder.
Each block contains an expert knowledge extractor, a dynamic gating network, and a transformer block.
For the $i$-th block, the input feature is denoted as $\mathbf{X}^{i} \in \mathbb{R}^{C \times H \times W}$ and we take the CLIP base encoder feature $\mathbf{X} \in \mathbb{R}^{C \times H \times W}$ as the input feature $\mathbf{X}^{1}$ of the first block.
We use $\mathbf{G}$ to indicate the indices of chosen $K$ experts.
The expert feature set is $\{\mathcal{F}_j|j \in \mathbf{G}\}$.
The final output feature of $L$ adapter blocks is $\mathbf{X}^{L+1}$. Additionally, we apply two residual blocks~\cite{resnet} with an average pooling to $\mathbf{X}^{L+1}$ to obtain a coarser image feature $\mathbf{X}^{L+1}_{out} \in \mathbb{R}^{C \times \frac{H}{2} \times \frac{W}{2}}$, which is further connected to the LLM text embedding space by an MLP layer.

\textbf{Text Encoder.}
We introduce a pre-trained BERT as the text encoder to extract language context information from the user's instruction. 
We take the [CLS] token from the output of the text encoder as the text token $\mathbf{X}_{T}  \in \mathbb{R}^{C_{T}}$.
It is worth noting that all the adapter blocks share the same text token.

\textbf{Expert Knowledge Extractor.}
We adopt $N$ cross-attention layers as the expert knowledge extractor to achieve efficient knowledge extraction.
Note that only the expert features $\{\mathcal{F}_j|j \in \mathbf{G}\}$ and their corresponding cross-attention layers are involved in the extraction.
For each selected expert feature $\mathcal{F}_{j} \in \mathbb{R}^{C_{j} \times H_{j} \times W_{j}}$, we first align its resolution to $\mathbf{X}^{i}$ with bilinear interpolation:
\begin{equation}
  \hat{\mathcal{F}}_{j} = {\rm Interpolate}({\mathcal{F}}_{j}, H, W).
\end{equation}
For the $i$-th MoV-Adapter block and the $j$-th cross-attention layer, we take input feature $\mathbf{X}^{i}$ as query, and the aligned expert feature $\hat{\mathcal{F}}_{j}$ as the key and value:
\begin{equation}
  \mathbf{Y}_{j}^{i} = \mathbf{X}^{i} + {\rm Attention}(\mathbf{X}^{i}, \hat{\mathcal{F}}_{j}).
\end{equation}

\textbf{Dynamic Gating Network.}
We employ a dynamic gating network to contribute to a fine-grained knowledge integration process for the conditional representation $\{\mathbf{Y}_{j}^{i}|j \in \mathbf{G}\}$.
It is implemented with the softmax over the logits of an MLP layer, processing multimodal representation to generate expert-wise soft weight $\mathbf{P}^{i} \in \mathbb{R}^{K}$ for the output of each cross-attention layer in the extractor.
Specifically, the input to the gating network is the concatenated vector of a visual token $\mathbf{X}^{i}_{V} \in \mathbb{R}^{C}$ and the text token $\mathbf{X}_{T}  \in \mathbb{R}^{C_{T}}$.
We obtain $\mathbf{X}^{i}_{V}$ with a global average pooling operation to $\mathbf{X}^{i}$.
Then we concatenate them to compute the gating weights and the expert-wise outputs by computing the weighted sum:
\begin{equation}
  \hat{\mathbf{X}}^{i} = \sum_{j\in \mathbf{G}} \mathbf{Y}_{j}^{i} \cdot \mathbf{P}^{i}_{j},
\end{equation}
where $\mathbf{P}^{i}_{j} \in (0, 1)$ is the soft weight for the $j$-th expert in the $i$-th block.

\textbf{Transformer Block.}
The transformer block in the adapter block follows the vanilla design, consisting of a self-attention layer and an FFN layer.
Taking the fused visual representation $\hat{\mathbf{X}}^{i}$, its output will serve as the input feature $\mathbf{X}^{i+1}$ for the next adapter block.

\begin{figure*}[tp]
    \centering
    \includegraphics[width=0.8\textwidth]{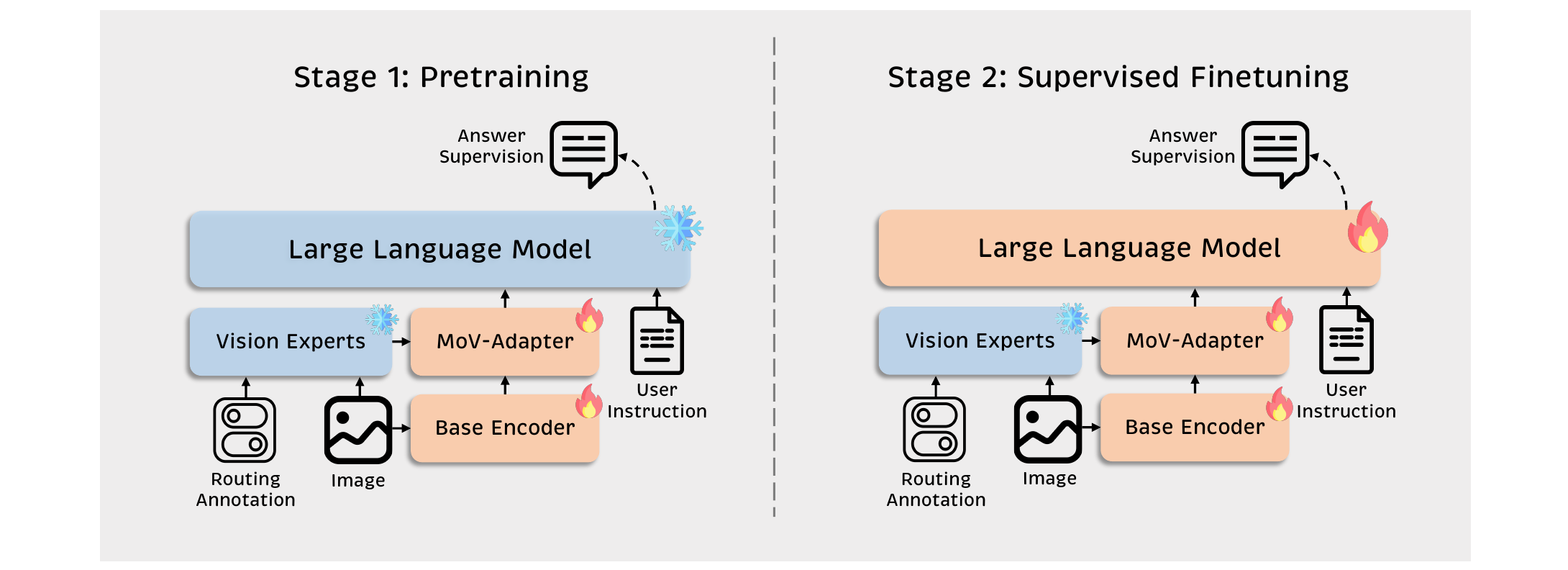}
    \caption{\textbf{The training strategy of MoVA.} We enhance the task-specific knowledge extraction capacity in the first stage. Then, we excite model multimodal capacities in the last stage. 
    }
    \label{fig:training}
\end{figure*}

\subsection{Training Paradigm}
\label{sec:training}
As shown in Figure~\ref{fig:training}, the training process of MoVA consists of pretraining and supervised finetuning.

\textbf{Pretraining.}
To improve multimodal generalization, we first construct 15M visual instruction samples across diverse domains as the training data: 
\textbf{(i)} Image caption data that covers 4M randomly selected samples from DataComp-1B~\cite{datacomp}, ShareGPT4V-PT~\cite{sharegpt4v}, and ALLaVA-4V~\cite{allava}. 
\textbf{(ii)} Visual grounding and localization dataset that encompasses Objects365~\cite{objects365},  RefCOCO~\cite{refcoco}, VisualGenome~\cite{visualgenome}, PointQA~\cite{pointqa}, and Flickr30K~\cite{flickr30k}.
\textbf{(iii)} Chart understanding data that includes MMC-Instruction~\cite{mmc}, Chart2Text~\cite{chart2text}, DVQA~\cite{dvqa}, and SciGraphQA~\cite{scigraphqa}.
\textbf{(iv)} Text recognition and document parsing data that covers LLaVAR-PT~\cite{llavar} and 3M English document images from Common Crawl~\footnote{https://commoncrawl.org}.
\textbf{(v)} LLaVA-Med~\cite{llava_med} for biomedical image understanding.
During the pretraining phase, we only optimize the MoV-Adapter along with the base vision encoder while preserving the capabilities of the initial large language model.
Meanwhile, we leverage the routing annotations generated via the method proposed in Section~\ref{sec:routing_data} to choose experts and ignore representations from irrelevant ones during training.

\textbf{Supervised Finetuning.}
We utilize high-quality visual instruction tuning data that build upon LLaVA-665K~\cite{llava15} for finetuning.
Additionally, we integrate several visual question answering datasets across various domains, such as DocVQA~\cite{docvqa}, ChartQA~\cite{chartqa}, InfographicVQA~\cite{infographicvqa}, AI2D~\cite{ai2d}, ST-VQA~\cite{stvqa}, TextVQA~\cite{textvqa}, SynthDoG-en~\cite{donut}, Geometry3K~\cite{geo3k}, PGPS9K~\cite{pgps9k}, Geo170K~\cite{gllava}, RefCOCO, LLaVA-Med, VQA-RAD~\cite{vqa_rad}, and SLAKE~\cite{slake}.
We also encompass equivalent comprehensive captions~\cite{sharegpt4v,allava,laion4v,textocr4v} generated by the advanced GPT4-V~\cite{gpt4} for improved world knowledge.
Apart from the above instruction tuning data, we convert the selected 2K routing annotations to instructions and incorporate them into the training data.
In the supervised fine-tuning stage, only task-specific vision experts are frozen and we jointly optimize other components.
The objective of supervised fine-tuning is to align the visual representation and the embedding of LLM, boosting its visual instruction-following capabilities.

\section{Experiments}
\label{sec:experiments}

\subsection{Implementation Details}
As mentioned in Section~\ref{sec:training}, our training pipeline consists of two stages. 
In the pretraining stage, we use the AdamW optimizer with an initial learning rate of $2$$\times$$10^{-4}$, a batch size of 1024, and train the model for 1 epoch.
We jointly finetune the weights of all components except additional vision experts with a batch size of 128 and an initial learning rate of $2$$\times$$10^{-5}$ during supervised finetuning.
We use 3 transformer blocks ($L$$=$$3$) in the MoV-Adapter and its hidden dimension is 1024, which is consistent with the base vision encoder CLIP.
The input resolution of the base vision encoder is 672$\times$672.
Two residual blocks with an average pooling are employed in the MoV-Adapter to reduce the number of output image tokens from 2304 to 576.
For the experiment performed in Table~\ref{tab:intro_exp}, we follow the default setting of LLaVA-1.5 but incorporate several additional datasets, including DocVQA~\cite{docvqa}, ChartQA~\cite{chartqa}, RefCOCO~\cite{refcoco}, LLaVA-Med~\cite{llava_med}, VQA-RAD~\cite{vqa_rad}, and SLAKE~\cite{slake}.
More details about vision experts, ablations, and analysis are available in Appendix~\ref{sec:appendix_vision} and~\ref{sec:appendix_exp}.

\begin{table}[t]
    \centering
    \caption{\textbf{Performance comparison with current state-of-the-art frameworks on popular MLLM benchmarks.} PT and SFT indicate the number of multimodal training samples in pretraining and finetuning stage. \#IMG means the number of image tokens processed by LLM.}
    \label{tab:main_exp}
    \setlength\tabcolsep{2pt}
    \resizebox{0.95\textwidth}{!}{
        \begin{tabular}{l|l|ccc|ccccccc}
        \toprule
        Model & LLM & PT & SFT & \#IMG & MME & MMB & MMB$^{{\rm{CN}}}$ & QBench & MathVista & MathVerse & POPE \\
        \midrule
        \multicolumn{12}{c}{\it Proprietary MLLMs} \\
        \midrule
        Qwen-VL-Plus~\cite{qwenvl} & \multicolumn{1}{c|}{--} & -- & -- & -- & -- & 66.2 & 68.0 & 66.0 & 43.3 & 11.8 & -- \\   
        Qwen-VL-Max~\cite{qwenvl} & \multicolumn{1}{c|}{--} & -- & -- & -- & -- & 77.6 & 75.1 & 73.6 & 51.0 & 24.8 & -- \\           
        Gemini-Pro~\cite{gemini} & \multicolumn{1}{c|}{--} & -- & -- & -- & -- & 73.6 & 74.3 & 68.2 & 45.2 & 22.3 & -- \\      
        GPT-4V~\cite{gpt4} & \multicolumn{1}{c|}{--} & -- & -- & -- & -- & 75.8 & 73.9 & 74.5 & 49.9 & 38.3 & -- \\          
        \midrule
        \multicolumn{12}{c}{\it Open-source MLLMs} \\
        \midrule
        Qwen-VL~\cite{qwenvl} & Qwen-7B & 1.4B & 50M & 256 & -- & 38.2 & 7.4 & 59.4 & -- & -- & -- \\
        Qwen-VL-Chat~\cite{qwenvl} & Qwen-7B & 1.4B & 50M & 256 & 1488/361 & 60.6 & 56.7 & -- & -- & -- & -- \\
        LLaVA-1.5~\cite{llava15} & Vicuna-7B & 558K & 665K & 576 & 1511/316 & 64.3 & 58.3 & 58.7 & -- & 14.3 & 85.9 \\
        LLaVA-1.5~\cite{llava15} & Vicuna-13B & 558K & 665K & 576 & 1531/295 & 67.7 & 63.6 & 62.1 & 27.6 & 17.0 & 85.9  \\
        mPLUG-Owl2~\cite{mplugowl2} & LLaMA2-7B & 348M & 1.2M & 64 & 1450/-- & 64.5 & -- & 62.9 & -- & 4.6 & 85.8 \\
        SPHINX-2k~\cite{sphinx} & Vicuna-13B & 115M & -- & 2880 & 1471/327 & 65.9 & 57.9 & -- & 27.8 & -- & 87.2 \\   
        LLaVA-NeXT~\cite{llavanext} & Vicuna-7B & 558K & 760K & 2880 & 1519/332 & 67.4 & 60.6 & -- & 34.6 & -- & 86.5 \\
        LLaVA-NeXT~\cite{llavanext} & Hermes-Yi-34B & 558K & 760K & 2880 & \textbf{1631}/397 & 79.3 & 79.0 & -- & \textbf{46.5} & 23.4 & 87.7  \\    
        \midrule        
        \cellcolor{gray!15}\textbf{MoVA} & \cellcolor{gray!15}Vicuna-7B & \cellcolor{gray!15}15M & \cellcolor{gray!15}1.6M & \cellcolor{gray!15}576 & \cellcolor{gray!15}1562/371 & \cellcolor{gray!15}70.4 & \cellcolor{gray!15}63.7 & \cellcolor{gray!15}69.3 & \cellcolor{gray!15}37.6 & \cellcolor{gray!15}19.7 & \cellcolor{gray!15}88.6 \\        
        \cellcolor{gray!15}\textbf{MoVA} & \cellcolor{gray!15}Llama3-8B & \cellcolor{gray!15}15M & \cellcolor{gray!15}1.6M & \cellcolor{gray!15}576 & \cellcolor{gray!15}1596/348 & \cellcolor{gray!15}75.3 & \cellcolor{gray!15}67.7 & \cellcolor{gray!15}\textbf{70.8} & \cellcolor{gray!15}37.7 & \cellcolor{gray!15}21.4 & \cellcolor{gray!15}\textbf{89.3} \\          
        \cellcolor{gray!15}\textbf{MoVA} & \cellcolor{gray!15}Hermes-Yi-34B & \cellcolor{gray!15}15M & \cellcolor{gray!15}1.6M & \cellcolor{gray!15}576 & \cellcolor{gray!15}1603/\textbf{455} & \cellcolor{gray!15}\textbf{81.3} & \cellcolor{gray!15}\textbf{79.0} & \cellcolor{gray!15}70.7 & \cellcolor{gray!15}44.3 & \cellcolor{gray!15}\textbf{23.7} &
        \cellcolor{gray!15}88.3 \\        
        \bottomrule
        \end{tabular}   
}
\end{table}

\subsection{MLLM Benchmarks}
We empirically analyze the multimodal capacity and generalization ability of MoVA on a wide range of challenging MLLM benchmarks in Table~\ref{tab:main_exp}.
This comprehensive assessment is conducted on MME~\cite{mme}, MMBench~\cite{mmbench}, QBench~\cite{qbench}, MathVista~\cite{mathvista}, MathVerse~\cite{mathverse}, and POPE~\cite{pope}.
Compared to other open-source MLLMs with similar model complexity, MoVA with Vicuna-7B achieves the best performance across 7 MLLM benchmarks while offering a more favorable balance between training efficiency and performance.
For instance, MoVA-7B surpasses the recent state-of-the-art LLaVA-NeXT-7B~\cite{llavanext} with a dynamic high resolution design, processing only 20\% image tokens.
Furthermore, we adopt Hermes-Yi-34B~\cite{yi} as the LLM to validate the scaling property of MoVA.
As depicted in Table~\ref{tab:main_exp}, the performance of MoVA-34B is on par with popular proprietary MLLMs (\textit{e.g.}, Gemini-Pro~\cite{gemini}) and outperforms Qwen-VL-Plus~\cite{qwenvl} on 5 MLLM benchmarks.
For example, MoVA establishes new records on MMBench and MMBench-CN, even surpassing the GPT-4V~\cite{gpt4}.
These results suggest that the ensemble of vision experts with adaptive expert routing can serve as an effective dimension for MLLM model scaling.

\subsection{Visual Question Answering}
\begin{table*}[t]
\centering
\caption{\textbf{Performance comparison on VQA benchmarks}. We present the number of model parameters of each MLLM for a clear complexity comparison.
* denotes zero-shot evaluation.
}
\label{tab:vqa}
\setlength\tabcolsep{2pt}
\resizebox{0.9\textwidth}{!}{%
    \begin{tabular}{l|l|c|ccc|cccc}
    \toprule
    \multirow{2}{*}{Model} & \multirow{2}{*}{LLM} & \multirow{2}{*}{Params} & \multicolumn{3}{c|}{General VQA} & \multicolumn{4}{c}{Text-oriented VQA} \\
      &  & & VQA$^{\rm{v2}}$ & GQA & SQA$^{\rm{I}}$ & TextVQA  & ChartQA & DocVQA & AI2D \\ 
    \midrule
    \multicolumn{10}{c}{\it Generalist models} \\
    \midrule
    Qwen-VL~\cite{qwenvl} & Qwen-7B & 10B & 79.5 & 59.3 & 67.1$^{*}$ & 63.8 & 65.7 & 65.1 & 62.3 \\ 
    Qwen-VL-Chat~\cite{qwenvl} & Qwen-7B & 10B & 78.2 & 57.5 & 68.2$^{*}$ & 61.5 & 66.3 & 62.6 & 57.7 \\    
    LLaVA-1.5~\cite{llava15} & Vicuna-7B & 7B & 78.5 & 62.0 & 66.8$^{*}$ & 58.2$^{*}$ & -- & -- & -- \\
    LLaVA-1.5~\cite{llava15} & Vicuna-13B & 7B & 80.0 & 63.3 & 71.6$^{*}$ & 61.3$^{*}$ & -- & -- & -- \\    
    SPHINX-2k~\cite{sphinx} & Vicuna-13B & 16B & 80.7 & 63.1 & 70.6$^{*}$ & 61.2 & -- & -- & 65.1 \\
    Vary-base~\cite{vary} & Qwen-7B & 7B & -- & -- & -- & -- & 65.3 & 76.3 & -- \\
    CogAgent~\cite{cogagent} & Vicuna-7B & 18B & 83.7 & -- & -- & 76.1 & 68.4 & 81.6 & -- \\
    \midrule
    \multicolumn{10}{c}{\it Specialist models} \\
    \midrule
    Pix2Struct-Large~\cite{pix2struct} & \multicolumn{1}{c|}{--} & 1.3B & -- & -- & -- & -- & 58.6 & 76.6 & 42.1 \\
    PALI-X-55B~\cite{palix} & \multicolumn{1}{c|}{--} & 55B & \textbf{86.0} & -- & -- & 71.4 & 70.9 & 80.0 & 81.2 \\ 
    \midrule
    \cellcolor{gray!15}\textbf{MoVA} & \cellcolor{gray!15}Vicuna-7B & \cellcolor{gray!15}10B & \cellcolor{gray!15}83.5 & \cellcolor{gray!15}64.8 & \cellcolor{gray!15}74.4$^{*}$ & \cellcolor{gray!15}76.4 & \cellcolor{gray!15}68.3 & \cellcolor{gray!15}81.3 & \cellcolor{gray!15}74.9 \\    
    \cellcolor{gray!15}\textbf{MoVA} & \cellcolor{gray!15}Llama3-8B & \cellcolor{gray!15}11B & \cellcolor{gray!15}83.5 & \cellcolor{gray!15}65.2 & \cellcolor{gray!15}74.7$^{*}$ & \cellcolor{gray!15}77.1 & \cellcolor{gray!15}70.5 & \cellcolor{gray!15}83.4 & \cellcolor{gray!15}77.0 \\       
    \cellcolor{gray!15}\textbf{MoVA} & \cellcolor{gray!15}Hermes-Yi-34B & \cellcolor{gray!15}38B & \cellcolor{gray!15}82.3 & \cellcolor{gray!15}63.9 & \cellcolor{gray!15}\textbf{79.0}$^{*}$ & \cellcolor{gray!15}\textbf{77.8} & \cellcolor{gray!15}\textbf{73.8} & \cellcolor{gray!15}\textbf{84.2} & \cellcolor{gray!15}\textbf{83.0} \\  
    \bottomrule
    \end{tabular}
}
\end{table*}
The evaluation results on VQA benchmarks are outlined in Table~\ref{tab:vqa}.
We divide these benchmarks into general VQA benchmarks~\cite{vqav2,gqa,scienceqa} and text-oriented VQA benchmarks~\cite{textvqa,chartqa,docvqa,ai2d}.
Thanks to the dynamic and efficient task-specific knowledge extraction, MoVA achieves state-of-the-art performances across diverse VQA benchmarks.
For general VQA benchmarks, MoVA-7B outperforms SPHINX-2k~\cite{internvl} equipped with Vicuna-13B on VQAv2~\cite{vqav2} and GQA by 4.2\% and 1.9\%, respectively.
Besides, MoVA shows its proficiency in text recognition in various scenarios, encompassing scene text, chart, document, and diagram.
For instance, MoVA-7B catches up to the current state-of-the-art generalist CogAgent~\cite{cogagent} with 18 billion parameters on these text-oriented benchmarks with smaller model size.
The MoVA model with 38B parameters even surpasses the well-established specialist model PALI-X-55B~\cite{palix} by clear margins.
These outstanding performances demonstrate MoVA’s robust generalization capabilities across diverse domains.

\subsection{Visual Grounding}
\begin{table*}[t]
\centering
\caption{\textbf{Performance comparison (Acc@0.5) on RefCOCO REC task.}
Specialists are specifically designed for the grounding task or finetuned on RefCOCO data.
}
\label{tab:rec}
\setlength\tabcolsep{3pt}
\resizebox{0.9\textwidth}{!}{%
    \begin{tabular}{l|l|ccc|ccc|cc}
    \toprule
    \multirow{2}{*}{Type} & \multirow{2}{*}{Model}  & \multicolumn{3}{c|}{RefCOCO} & \multicolumn{3}{c|}{RefCOCO+} & \multicolumn{2}{c}{RefCOCOg} \\
     &  & val & test-A & test-B & val & test-A & test-B & val & test \\
    \midrule
    \multirow{6}{*}{Generalist}
    & Shikra-13B~\cite{shikra} & 87.83 & 91.11 & 81.81 & 82.89 & 87.79 & 74.41 & 82.64 & 83.16 \\    
    & Ferret-13B~\cite{ferret}  & 89.48 & 92.41 & 84.36 & 82.81 & 88.14 & 75.17 & 85.83 & 86.34 \\
    & Qwen-VL~\cite{qwenvl} & 89.36 & 92.26 & 85.34 & 83.12 & 88.25 & 77.21 & 85.58 & 85.48 \\
    & SPHINX-2k~\cite{sphinx} & 91.10 & 92.88 & 87.07 & 85.51 & 90.62 & 80.45 & 88.07 & 88.65 \\  
    & \cellcolor{gray!15}\textbf{MoVA-7B} & \cellcolor{gray!15}92.55 & \cellcolor{gray!15}94.50 & \cellcolor{gray!15}88.81 & \cellcolor{gray!15}87.70 & \cellcolor{gray!15}92.05 & \cellcolor{gray!15}82.94 & \cellcolor{gray!15}89.28 & \cellcolor{gray!15}89.70 \\
    & \cellcolor{gray!15}\textbf{MoVA-8B} & 
    \cellcolor{gray!15}92.18 & \cellcolor{gray!15}\textbf{94.75} & \cellcolor{gray!15}88.24 & \cellcolor{gray!15}88.45 & \cellcolor{gray!15}92.21 & \cellcolor{gray!15}82.82 & \cellcolor{gray!15}90.05 & \cellcolor{gray!15}90.23 \\    
    & \cellcolor{gray!15}\textbf{MoVA-34B} & \cellcolor{gray!15}\textbf{93.38} & \cellcolor{gray!15}94.66 & \cellcolor{gray!15}\textbf{90.58} & \cellcolor{gray!15}\textbf{89.64} & \cellcolor{gray!15}\textbf{92.53} & \cellcolor{gray!15}\textbf{84.03} & \cellcolor{gray!15}\textbf{91.09} & \cellcolor{gray!15}\textbf{90.78} \\    
    \midrule
    \multirow{2}{*}{Specialist}
     & G-DINO-L~\cite{groundingdino}     & 90.56 & 93.19 & 88.24 & 82.75 & 88.95 & 75.92 & 86.13 & 87.02 \\
     & UNINEXT-H~\cite{uninext}    & 92.64 & 94.33 & 91.46 & 85.24 & 89.63 & 79.79 & 88.73 & 89.37 \\
    \bottomrule
    \end{tabular}%
}
\end{table*}
We conduct experiments on Referring Expression Comprehension (REC) benchmarks~\cite{refcoco} to evaluate the visual grounding ability of MoVA.
The results are presented in Table\ref{tab:rec}.
The performance of MoVA-7B is on par with the state-of-the-art specialist models that are elaborately designed for grounding tasks.
For example, MoVA-7B achieves a score of 90.22\% on RefCOCO+ val, which is 2.46\% higher than the score of UNINEXT-H~\cite{uninext}.
Our largest model MoVA-34B further pushes the performance bound of visual grounding on these benchmarks.
These impressive results demonstrate MoVA's remarkable visual grounding capacity.

\subsection{Medical Visual Question Answering}
This experiment is conducted on popular medical VQA benchmarks VQA-RAD and SLAKE.
We directly leverage the medical VQA evaluation metric adopted by LLaVA-Med.
Each sample of VQA-RAD and SLAKE is observed only once during the training process of MoVA and LLaVA-1.5.
For a fair comparison, we compare MoVA with the LLaVA-Med variant that is finetuned with only 1 epoch on the benchmark.
The performance of the LLaVA-Med specialist that is fully finetuned on downstream tasks is also reported. 
As presented in Table~\ref{tab:ablation_med}, MoVA-7B consistently yields higher scores than LLaVA-Med and LLaVA-1.5, exhibiting its medical visual chat ability.


\begin{table*}[t]
    \begin{minipage}[t]{0.34\linewidth}
        \centering
        \caption{Comparisons on the biomedical VQA datasets.
        }
        \resizebox{\textwidth}{!}{
            \begin{tabular}{@{}c|cc|cc@{}}
            \toprule
            \multirow{2}{*}{Model} & \multicolumn{2}{c|}{VQA-RAD} & \multicolumn{2}{c}{SLAKE} \\
             & Open & Close & Open & Close \\
            \midrule
            LLaVA-Med & 28.6 & 56.3 & 70.6 & 54.6 \\    
            LLaVA-1.5 & 35.3 & 68.9 & 73.1 & 63.7 \\
            \cellcolor{gray!15}\textbf{MoVA} & \cellcolor{gray!15}\textbf{38.3} & \cellcolor{gray!15}\textbf{68.9} & \cellcolor{gray!15}\textbf{78.2} & \cellcolor{gray!15}\textbf{68.8} \\
            \midrule                    
            \color{dt}LLaVA-Med (ft) & \color{dt}61.5 & \color{dt}84.2 & \color{dt}83.1 & \color{dt}85.3 \\    
            \bottomrule
            \end{tabular}
        }
        \label{tab:ablation_med}
    \end{minipage}
    \hspace{0.2cm}
    \begin{minipage}[t]{0.36\linewidth}
        \centering
        \caption{
        Results of component-wise ablation studies.
        }
        \resizebox{\textwidth}{!}{
            \begin{tabular}{@{}c|ccc@{}}
            \toprule
            Design & GQA & ChartQA & DocVQA \\
            \midrule
            \cellcolor{gray!15}\textbf{MoVA} & \cellcolor{gray!15}\textbf{64.8} & \cellcolor{gray!15}\textbf{68.3} & \cellcolor{gray!15}\textbf{81.3} \\
            \midrule
            Random routing & 63.1 & 60.4 & 71.6 \\
            w/o routing & 63.4 & 62.5 & 73.7 \\
            \midrule
            w/o MoV-Adapter & 62.7 & 65.2 & 77.1 \\
            \bottomrule
            \end{tabular}
        }
        \label{tab:component}
    \end{minipage}
    \hspace{0.2cm}
    \begin{minipage}[t]{0.24\linewidth}
        \centering
        \caption{Results of $K$ varying from 1 to 3. 
        }
        \resizebox{\textwidth}{!}{
            \begin{tabular}{@{}c|ccccc@{}}
            \toprule
            $K$ & GQA & ChartQA \\
            \midrule
            \cellcolor{gray!15}\textbf{Dynamic} & \cellcolor{gray!15}\textbf{64.8} & \cellcolor{gray!15}\textbf{68.3} \\
            \midrule
            1 & 64.0 & 64.9 \\
            2 & 63.5 & 66.7 \\
            3 & 63.2 & 67.4 \\
            \bottomrule
            \end{tabular}
        }
        \label{tab:ablation_K}
    \end{minipage}    
\end{table*}
\begin{table*}[t]
    \begin{minipage}[t]{0.33\linewidth}
        \centering
        \caption{Comparisons of expert routing criteria.
        }
        \resizebox{\textwidth}{!}{
            \begin{tabular}{@{}c|c|cccc@{}}
            \toprule
            Design & \#Models & POPE & GQA & ChartQA \\
            \midrule
            \cellcolor{gray!15}\textbf{Separate} & \cellcolor{gray!15}4 & \cellcolor{gray!15}88.6 & \cellcolor{gray!15}\textbf{64.8} & \cellcolor{gray!15}68.3 \\ 
            \midrule
            Combination & \textbf{14} & \textbf{88.9} & 64.6 & \textbf{68.7} \\
            \bottomrule
            \end{tabular}
        }
        \label{tab:ablation_combination}
    \end{minipage}
    \hspace{0.2cm}
    \begin{minipage}[t]{0.27\linewidth}
        \centering
        \caption{Open-world expert routing results.
        }
        \resizebox{\textwidth}{!}{
            \begin{tabular}{@{}cc|c@{}}
            \toprule
            Design & \#Samples & Accuracy \\
            \midrule
            \cellcolor{gray!15}\textbf{MoVA} & \cellcolor{gray!15}2K & \cellcolor{gray!15}\textbf{92.5\%} \\
            \midrule
            MoVA & \textbf{5K} & 12.5\% \\
            w/o Augmentation & 2K & 0\% \\
            MLP classifier & 2K & 0\% \\
            \bottomrule
            \end{tabular}
        }
        \label{tab:ablation_open}
    \end{minipage}        
    \hspace{0.2cm}
    \begin{minipage}[t]{0.32\linewidth}
        \centering
        \caption{Performance of various MoV-Adapter variants.
        }
        \resizebox{\textwidth}{!}{
            \begin{tabular}{@{}c|ccccccc@{}}
            \toprule
            Design & MME$^{\rm{P}}$ & MMB & POPE & GQA \\
            \midrule
            \cellcolor{gray!15}\textbf{MoVA} & \cellcolor{gray!15}1562 & \cellcolor{gray!15}\textbf{70.4} & \cellcolor{gray!15}\textbf{88.6} & \cellcolor{gray!15}\textbf{64.8} \\
            \midrule
            2 blocks & 1526 & 70.1 & 87.9 & 63.9  \\
            4 blocks & \textbf{1578} & 69.4 & 88.3 & 64.5 \\
            \midrule
            Uniform gating & 1521 & 69.1 & 87.5 & 64.1 \\
            \bottomrule
            \end{tabular}
        }
        \label{tab:ablation_mov}
    \end{minipage}    
\end{table*}


\subsection{Ablation Study}
\textbf{Component-wise analysis.}
As presented in Table~\ref{tab:component}, we perform an ablation to thoroughly delve into the effect of each component.
First, we try to replace the context-aware routing with random routing.
Without task-relevant vision experts, the performance drops by a large margin, especially on the text-oriented VQA benchmarks.
Removing context-aware routing to leverage all vision experts also leads to similar results.
It proves that both these modifications introduce biased information from irrelevant vision experts due to the removal of context-aware routing.
Then, we ablate the effectiveness of the MoV-Adapter by replacing it with simple linear layers.
The removal of fine-grained expert feature fusion downgrades performance across all datasets.
These results delineate that each component in MoVA can consistently yield significant gains.

\textbf{Number of activated experts.}
In the context-aware routing phase, the number of activated experts $K$ is dynamic.
We compare such a data-dependent design with other variations of constant $K$ in this experiment.
As presented in Table~\ref{tab:ablation_K}, the overall performance of dynamic $K$ consistently outperforms other models with constant $K$.
This reveals this dynamic implementation can fully exploit the task-specific knowledge of relevant experts while avoiding the incorporation of biased information.

\textbf{Criteria for choosing better experts.}
To reduce the costs, our method only adopts $N \choose 1$ models with $N$ various encoders to identify the better vision expert \textit{separately}.
However, we do not explicitly consider the combination of the chosen vision experts.
In this experiment, we compare our method with another strategy that considers vision encoders combination and enumerates $\sum_{i=1}^{3}{N \choose i}$ models for routing data construction.
Specifically, we set $N=4$ and employ a smaller model pool of DINOv2, Co-DETR, Pix2Struct, and Deplot to reduce training costs.
As shown in Table~\ref{tab:ablation_combination}, our method achieves comparable performance while requiring much less models for data construction.

\textbf{Expert routing in open scenarios.}
We develop 105 human-verified testing samples that should be answered using novel experts for the expert routing task.
These novel experts encompass 7 vision models~\cite{hop,donut,dino_det,groundingdino,layoutlmv3,resnet,sit} on various computer vision tasks and each expert corresponds to 15 evaluation samples.
We manually check the correctness of the expert routing result.
As presented in Table~\ref{tab:ablation_open}, a lightweight network, such as a MLP classifier fails to generalize to this open-world setting.
Besides, increasing the routing training samples and removing data augmentations also lead to severe performance degradation.
The results demonstrate our coarse-grained context-aware routing preserves the generalization ability for expert routing in open scenarios.

\textbf{Adapter Design.}
In this section, we conduct ablation studies on the design of the MoV-Adapter.
As presented in Table \ref{tab:ablation_mov}, we compared the impact of using 2, 3, and 4 adapter blocks on the model's performance.
We observed that the baseline with 3 blocks can achieve better performance than other settings.
Then, we substituted our multimodal gating for uniform gating to investigate its effectiveness.
Each of the experts is assigned the same soft weight in the uniform gating.
We find uniform gating brings consistent performance drops in the test benchmarks.
It indicates that the lack of the dynamic soft-weight harms the overall performance since it fails to perform precise knowledge extraction.

\textbf{Inference Analysis.}
As illustrated in Figure~\ref{fig:framework}, MoVA consists of two stages: coarse-grained context-ware expert routing and fine-grained expert fusion with MoV-Adapter. This two-stage inference pipeline can be further broken down into five steps: \textbf{(i)} Data preprocessing. We first process the input image with image processors and convert the input text into a token sequence with the LLM tokenizer. \textbf{(ii)} Base encoder forward. We extract the base image feature using the base CLIP encoder. Note that we only run the base encoder once since its output feature can be preserved and reused in the fourth step. \textbf{(iii)} LLM routing generation. We compress the base image features into 144 image tokens. The LLM generates a concise routing answer based on the compressed image feature and routing instruction. Vision experts and MoV-Adapter forward. \textbf{(iv)} According to the multimodal context and routing results generated in the previous step, we fuse vision features of the base encoder and activated experts in a coarse-to-fine manner. \textbf{(v)} LLM response generation. The LLM generates the final response given the fused vision features and user instructions. To investigate the inference efficiency of each step, we randomly select 200 images from the COCO val2017 dataset and adopt the common image caption instruction: \textit{Describe this image.} The temperature for generation is 0. The latency is measured using bfloat16 and flash-attention 2 on an A100 80G GPU. We present the average inference latency of each step and show the average sequence length of the routing output and final response. The inference latencies for each step are 0.19s, 0.05s, 0.14s, 0.07s, and 10.24s, respectively. 
The average length of the routing output is 3.24 tokens, while the average length of the final response is 405.06 tokens. Compared to the LLM response generation (Step 5), the LLM expert routing (Step 3) generates much fewer output tokens and its latency is negligible (0.14s \textit{v.s.} 10.24s). Therefore, our method does not bring significant inference costs.
\section{Conclusion}
In this paper, we reveal that the inherent bias of each vision encoder can diminish its generalization ability across other irrelevant domains by analyzing the performance of individual vision encoders versus the plain fusion of multiple encoders across various tasks.
To deal with the problem, we propose MoVA, a powerful MLLM composed of coarse-grained context-aware expert routing and fine-grained expert fusion with MoV-Adapter.
Based on multimodal context and model expertise, MoVA fully leverages representation from multiple context-relevant vision encoder experts flexibly and effectively while avoiding biased information brought by irrelevant experts.
MoVA can achieve significant performance gains over current state-of-the-art methods in a wide range of benchmarks.

\textbf{Limitations.} 
We acknowledge some limitations in our paper that require attention. 
One limitation is the hallucination, which refers to the generation of text that appears plausible or coherent but is factually incorrect and misleading.
This issue potentially presents in all powerful MLLMs.
Additionally, the performance may be affected by failure cases of the context-relevant vision experts, leading to potential degradation.
We plan to explore solutions for these limitations in future works.

\begin{ack}
The work was supported by the National Key R$\&$D Program of China under Grant 2021ZD0201300.
\end{ack}

\bibliographystyle{unsrt}
\bibliography{main}

\begin{thebibliography}{10}

\bibitem{blip2}
Junnan Li, Dongxu Li, Silvio Savarese, and Steven Hoi.
\newblock {BLIP}-2: Bootstrapping language-image pre-training with frozen image encoders and large language models.
\newblock In {\em International conference on machine learning}, pages 19730--19742. PMLR, 2023.

\bibitem{flamingo}
Jean-Baptiste Alayrac, Jeff Donahue, Pauline Luc, Antoine Miech, Iain Barr, Yana Hasson, Karel Lenc, Arthur Mensch, Katherine Millican, Malcolm Reynolds, et~al.
\newblock Flamingo: a visual language model for few-shot learning.
\newblock {\em Advances in Neural Information Processing Systems}, 35:23716--23736, 2022.

\bibitem{llava}
Haotian Liu, Chunyuan Li, Qingyang Wu, and Yong~Jae Lee.
\newblock Visual instruction tuning.
\newblock {\em Advances in neural information processing systems}, 36, 2024.

\bibitem{internvl}
Zhe Chen, Jiannan Wu, Wenhai Wang, Weijie Su, Guo Chen, Sen Xing, Zhong Muyan, Qinglong Zhang, Xizhou Zhu, Lewei Lu, et~al.
\newblock Internvl: Scaling up vision foundation models and aligning for generic visual-linguistic tasks.
\newblock {\em arXiv preprint arXiv:2312.14238}, 2023.

\bibitem{shikra}
Keqin Chen, Zhao Zhang, Weili Zeng, Richong Zhang, Feng Zhu, and Rui Zhao.
\newblock Shikra: Unleashing multimodal llm's referential dialogue magic.
\newblock {\em arXiv preprint arXiv:2306.15195}, 2023.

\bibitem{qwenvl}
Jinze Bai, Shuai Bai, Shusheng Yang, Shijie Wang, Sinan Tan, Peng Wang, Junyang Lin, Chang Zhou, and Jingren Zhou.
\newblock Qwen-vl: A frontier large vision-language model with versatile abilities.
\newblock {\em arXiv preprint arXiv:2308.12966}, 2023.

\bibitem{instructblip}
Wenliang Dai, Junnan Li, Dongxu Li, AnthonyMeng Huat, Junqi Zhao, Weisheng Wang, Boyang Li, Pascale Fung, and Steven Hoi.
\newblock Instructblip: Towards general-purpose vision-language models with instruction tuning.
\newblock {\em arXiv preprint arXiv:2305.06500}, 2023.

\bibitem{vicuna}
Wei-Lin Chiang, Zhuohan Li, Zi~Lin, Ying Sheng, Zhanghao Wu, Hao Zhang, Lianmin Zheng, Siyuan Zhuang, Yonghao Zhuang, Joseph~E. Gonzalez, Ion Stoica, and Eric~P. Xing.
\newblock Vicuna: An open-source chatbot impressing gpt-4 with 90\%* chatgpt quality, March 2023.

\bibitem{qwen}
Jinze Bai, Shuai Bai, Yunfei Chu, Zeyu Cui, Kai Dang, Xiaodong Deng, Yang Fan, Wenbin Ge, Yu~Han, Fei Huang, et~al.
\newblock Qwen technical report.
\newblock {\em arXiv preprint arXiv:2309.16609}, 2023.

\bibitem{zhang2024map}
Ge~Zhang, Scott Qu, Jiaheng Liu, Chenchen Zhang, Chenghua Lin, Chou~Leuang Yu, Danny Pan, Esther Cheng, Jie Liu, Qunshu Lin, et~al.
\newblock Map-neo: Highly capable and transparent bilingual large language model series.
\newblock {\em arXiv preprint arXiv:2405.19327}, 2024.

\bibitem{clip}
Alec Radford, Jong~Wook Kim, Chris Hallacy, Aditya Ramesh, Gabriel Goh, Sandhini Agarwal, Girish Sastry, Amanda Askell, Pamela Mishkin, Jack Clark, et~al.
\newblock Learning transferable visual models from natural language supervision.
\newblock In {\em International conference on machine learning}, pages 8748--8763. PMLR, 2021.

\bibitem{vary}
Haoran Wei, Lingyu Kong, Jinyue Chen, Liang Zhao, Zheng Ge, Jinrong Yang, Jianjian Sun, Chunrui Han, and Xiangyu Zhang.
\newblock Vary: Scaling up the vision vocabulary for large vision-language models.
\newblock {\em arXiv preprint arXiv:2312.06109}, 2023.

\bibitem{sphinx}
Ziyi Lin, Chris Liu, Renrui Zhang, Peng Gao, Longtian Qiu, Han Xiao, Han Qiu, Chen Lin, Wenqi Shao, Keqin Chen, et~al.
\newblock Sphinx: The joint mixing of weights, tasks, and visual embeddings for multi-modal large language models.
\newblock {\em arXiv preprint arXiv:2311.07575}, 2023.

\bibitem{mof}
Shengbang Tong, Zhuang Liu, Yuexiang Zhai, Yi~Ma, Yann LeCun, and Saining Xie.
\newblock Eyes wide shut? exploring the visual shortcomings of multimodal llms.
\newblock {\em arXiv preprint arXiv:2401.06209}, 2024.

\bibitem{dinov2}
Maxime Oquab, Timoth{\'e}e Darcet, Th{\'e}o Moutakanni, Huy Vo, Marc Szafraniec, Vasil Khalidov, Pierre Fernandez, Daniel Haziza, Francisco Massa, Alaaeldin El-Nouby, et~al.
\newblock Dinov2: Learning robust visual features without supervision.
\newblock {\em arXiv preprint arXiv:2304.07193}, 2023.

\bibitem{mmbench}
Yuan Liu, Haodong Duan, Yuanhan Zhang, Bo~Li, Songyang Zhang, Wangbo Zhao, Yike Yuan, Jiaqi Wang, Conghui He, Ziwei Liu, et~al.
\newblock Mmbench: Is your multi-modal model an all-around player?
\newblock {\em arXiv preprint arXiv:2307.06281}, 2023.

\bibitem{docvqa}
Minesh Mathew, Dimosthenis Karatzas, and CV~Jawahar.
\newblock Docvqa: A dataset for vqa on document images.
\newblock In {\em Proceedings of the IEEE/CVF winter conference on applications of computer vision}, pages 2200--2209, 2021.

\bibitem{chartqa}
Ahmed Masry, Do~Xuan Long, Jia~Qing Tan, Shafiq Joty, and Enamul Hoque.
\newblock Chartqa: A benchmark for question answering about charts with visual and logical reasoning.
\newblock {\em arXiv preprint arXiv:2203.10244}, 2022.

\bibitem{gqa}
Drew~A Hudson and Christopher~D Manning.
\newblock Gqa: A new dataset for real-world visual reasoning and compositional question answering.
\newblock In {\em Proceedings of the IEEE/CVF conference on computer vision and pattern recognition}, pages 6700--6709, 2019.

\bibitem{pope}
Yifan Li, Yifan Du, Kun Zhou, Jinpeng Wang, Wayne~Xin Zhao, and Ji-Rong Wen.
\newblock Evaluating object hallucination in large vision-language models.
\newblock {\em arXiv preprint arXiv:2305.10355}, 2023.

\bibitem{refcoco}
Licheng Yu, Patrick Poirson, Shan Yang, Alexander~C Berg, and Tamara~L Berg.
\newblock Modeling context in referring expressions.
\newblock In {\em Computer Vision--ECCV 2016: 14th European Conference, Amsterdam, The Netherlands, October 11-14, 2016, Proceedings, Part II 14}, pages 69--85. Springer, 2016.

\bibitem{slake}
Bo~Liu, Li-Ming Zhan, Li~Xu, Lin Ma, Yan Yang, and Xiao-Ming Wu.
\newblock Slake: A semantically-labeled knowledge-enhanced dataset for medical visual question answering.
\newblock In {\em 2021 IEEE 18th International Symposium on Biomedical Imaging (ISBI)}, pages 1650--1654. IEEE, 2021.

\bibitem{codetr}
Zhuofan Zong, Guanglu Song, and Yu~Liu.
\newblock Detrs with collaborative hybrid assignments training.
\newblock In {\em Proceedings of the IEEE/CVF international conference on computer vision}, pages 6748--6758, 2023.

\bibitem{sam}
Alexander Kirillov, Eric Mintun, Nikhila Ravi, Hanzi Mao, Chloe Rolland, Laura Gustafson, Tete Xiao, Spencer Whitehead, Alexander~C Berg, Wan-Yen Lo, et~al.
\newblock Segment anything.
\newblock {\em arXiv preprint arXiv:2304.02643}, 2023.

\bibitem{pix2struct}
Kenton Lee, Mandar Joshi, Iulia~Raluca Turc, Hexiang Hu, Fangyu Liu, Julian~Martin Eisenschlos, Urvashi Khandelwal, Peter Shaw, Ming-Wei Chang, and Kristina Toutanova.
\newblock Pix2struct: Screenshot parsing as pretraining for visual language understanding.
\newblock In {\em International Conference on Machine Learning}, pages 18893--18912. PMLR, 2023.

\bibitem{deplot}
Fangyu Liu, Julian~Martin Eisenschlos, Francesco Piccinno, Syrine Krichene, Chenxi Pang, Kenton Lee, Mandar Joshi, Wenhu Chen, Nigel Collier, and Yasemin Altun.
\newblock Deplot: One-shot visual language reasoning by plot-to-table translation.
\newblock {\em arXiv preprint arXiv:2212.10505}, 2022.

\bibitem{biomedclip}
Sheng Zhang, Yanbo Xu, Naoto Usuyama, Jaspreet Bagga, Robert Tinn, Sam Preston, Rajesh Rao, Mu~Wei, Naveen Valluri, Cliff Wong, et~al.
\newblock Large-scale domain-specific pretraining for biomedical vision-language processing.
\newblock {\em arXiv preprint arXiv:2303.00915}, 2023.

\bibitem{llava15}
Haotian Liu, Chunyuan Li, Yuheng Li, and Yong~Jae Lee.
\newblock Improved baselines with visual instruction tuning.
\newblock {\em arXiv preprint arXiv:2310.03744}, 2023.

\bibitem{toolllm}
Yujia Qin, Shihao Liang, Yining Ye, Kunlun Zhu, Lan Yan, Yaxi Lu, Yankai Lin, Xin Cong, Xiangru Tang, Bill Qian, et~al.
\newblock Toolllm: Facilitating large language models to master 16000+ real-world apis.
\newblock {\em arXiv preprint arXiv:2307.16789}, 2023.

\bibitem{lmdrive}
Hao Shao, Yuxuan Hu, Letian Wang, Steven~L Waslander, Yu~Liu, and Hongsheng Li.
\newblock Lmdrive: Closed-loop end-to-end driving with large language models.
\newblock {\em arXiv preprint arXiv:2312.07488}, 2023.

\bibitem{visualcot}
Hao Shao, Shengju Qian, Han Xiao, Guanglu Song, Zhuofan Zong, Letian Wang, Yu~Liu, and Hongsheng Li.
\newblock Visual cot: Unleashing chain-of-thought reasoning in multi-modal language models.
\newblock {\em arXiv preprint arXiv:2403.16999}, 2024.

\bibitem{jiao2024lumen}
Yang Jiao, Shaoxiang Chen, Zequn Jie, Jingjing Chen, Lin Ma, and Yu-Gang Jiang.
\newblock Lumen: Unleashing versatile vision-centric capabilities of large multimodal models.
\newblock {\em arXiv preprint arXiv:2403.07304}, 2024.

\bibitem{ma2024exploringrolelargelanguage}
Bingqi Ma, Zhuofan Zong, Guanglu Song, Hongsheng Li, and Yu~Liu.
\newblock Exploring the role of large language models in prompt encoding for diffusion models, 2024.

\bibitem{guo2023owl}
Hongcheng Guo, Jian Yang, Jiaheng Liu, Liqun Yang, Linzheng Chai, Jiaqi Bai, Junran Peng, Xiaorong Hu, Chao Chen, Dongfeng Zhang, et~al.
\newblock Owl: A large language model for it operations.
\newblock {\em arXiv preprint arXiv:2309.09298}, 2023.

\bibitem{zhou2024aligning}
Yiyang Zhou, Chenhang Cui, Rafael Rafailov, Chelsea Finn, and Huaxiu Yao.
\newblock Aligning modalities in vision large language models via preference fine-tuning.
\newblock {\em arXiv preprint arXiv:2402.11411}, 2024.

\bibitem{zhou2024calibrated}
Yiyang Zhou, Zhiyuan Fan, Dongjie Cheng, Sihan Yang, Zhaorun Chen, Chenhang Cui, Xiyao Wang, Yun Li, Linjun Zhang, and Huaxiu Yao.
\newblock Calibrated self-rewarding vision language models.
\newblock {\em arXiv preprint arXiv:2405.14622}, 2024.

\bibitem{jiang2024comat}
Dongzhi Jiang, Guanglu Song, Xiaoshi Wu, Renrui Zhang, Dazhong Shen, Zhuofan Zong, Yu~Liu, and Hongsheng Li.
\newblock Comat: Aligning text-to-image diffusion model with image-to-text concept matching.
\newblock {\em arXiv preprint arXiv:2404.03653}, 2024.

\bibitem{xia2024rule}
Peng Xia, Kangyu Zhu, Haoran Li, Hongtu Zhu, Yun Li, Gang Li, Linjun Zhang, and Huaxiu Yao.
\newblock Rule: Reliable multimodal rag for factuality in medical vision language models.
\newblock {\em arXiv preprint arXiv:2407.05131}, 2024.

\bibitem{xia2024mmed}
Peng Xia, Kangyu Zhu, Haoran Li, Tianze Wang, Weijia Shi, Sheng Wang, Linjun Zhang, James Zou, and Huaxiu Yao.
\newblock Mmed-rag: Versatile multimodal rag system for medical vision language models.
\newblock {\em arXiv preprint arXiv:2410.13085}, 2024.

\bibitem{yang2024boosting}
Zaiquan Yang, Yuhao Liu, Jiaying Lin, Gerhard Hancke, and Rynson~WH Lau.
\newblock Boosting weakly-supervised referring image segmentation via progressive comprehension.
\newblock {\em arXiv preprint arXiv:2410.01544}, 2024.

\bibitem{yue2024mmmu}
Xiang Yue, Yuansheng Ni, Kai Zhang, Tianyu Zheng, Ruoqi Liu, Ge~Zhang, Samuel Stevens, Dongfu Jiang, Weiming Ren, Yuxuan Sun, et~al.
\newblock Mmmu: A massive multi-discipline multimodal understanding and reasoning benchmark for expert agi.
\newblock In {\em Proceedings of the IEEE/CVF Conference on Computer Vision and Pattern Recognition}, pages 9556--9567, 2024.

\bibitem{li2024eyes}
Yian Li, Wentao Tian, Yang Jiao, Jingjing Chen, and Yu-Gang Jiang.
\newblock Eyes can deceive: Benchmarking counterfactual reasoning abilities of multi-modal large language models.
\newblock {\em arXiv preprint arXiv:2404.12966}, 2024.

\bibitem{zhang2024eventhallusion}
Jiacheng Zhang, Yang Jiao, Shaoxiang Chen, Jingjing Chen, and Yu-Gang Jiang.
\newblock Eventhallusion: Diagnosing event hallucinations in video llms.
\newblock {\em arXiv preprint arXiv:2409.16597}, 2024.

\bibitem{mmsearch}
Dongzhi Jiang, Renrui Zhang, Ziyu Guo, Yanmin Wu, Jiayi Lei, Pengshuo Qiu, Pan Lu, Zehui Chen, Guanglu Song, Peng Gao, et~al.
\newblock Mmsearch: Benchmarking the potential of large models as multi-modal search engines.
\newblock {\em arXiv preprint arXiv:2409.12959}, 2024.

\bibitem{xia2024cares}
Peng Xia, Ze~Chen, Juanxi Tian, Yangrui Gong, Ruibo Hou, Yue Xu, Zhenbang Wu, Zhiyuan Fan, Yiyang Zhou, Kangyu Zhu, et~al.
\newblock Cares: A comprehensive benchmark of trustworthiness in medical vision language models.
\newblock {\em arXiv preprint arXiv:2406.06007}, 2024.

\bibitem{xia2024mmie}
Peng Xia, Siwei Han, Shi Qiu, Yiyang Zhou, Zhaoyang Wang, Wenhao Zheng, Zhaorun Chen, Chenhang Cui, Mingyu Ding, Linjie Li, et~al.
\newblock Mmie: Massive multimodal interleaved comprehension benchmark for large vision-language models.
\newblock {\em arXiv preprint arXiv:2410.10139}, 2024.

\bibitem{wang2023rolellm}
Zekun~Moore Wang, Zhongyuan Peng, Haoran Que, Jiaheng Liu, Wangchunshu Zhou, Yuhan Wu, Hongcheng Guo, Ruitong Gan, Zehao Ni, Jian Yang, et~al.
\newblock Rolellm: Benchmarking, eliciting, and enhancing role-playing abilities of large language models.
\newblock {\em arXiv preprint arXiv:2310.00746}, 2023.

\bibitem{laion400m}
Christoph Schuhmann, Richard Vencu, Romain Beaumont, Robert Kaczmarczyk, Clayton Mullis, Aarush Katta, Theo Coombes, Jenia Jitsev, and Aran Komatsuzaki.
\newblock Laion-400m: Open dataset of clip-filtered 400 million image-text pairs.
\newblock {\em arXiv preprint arXiv:2111.02114}, 2021.

\bibitem{laion5b}
Christoph Schuhmann, Romain Beaumont, Richard Vencu, Cade Gordon, Ross Wightman, Mehdi Cherti, Theo Coombes, Aarush Katta, Clayton Mullis, Mitchell Wortsman, et~al.
\newblock Laion-5b: An open large-scale dataset for training next generation image-text models.
\newblock {\em Advances in Neural Information Processing Systems}, 35:25278--25294, 2022.

\bibitem{t2ibench}
Kaiyi Huang, Kaiyue Sun, Enze Xie, Zhenguo Li, and Xihui Liu.
\newblock T2i-compbench: A comprehensive benchmark for open-world compositional text-to-image generation.
\newblock {\em Advances in Neural Information Processing Systems}, 36, 2024.

\bibitem{sharegpt4v}
Lin Chen, Jisong Li, Xiaoyi Dong, Pan Zhang, Conghui He, Jiaqi Wang, Feng Zhao, and Dahua Lin.
\newblock Sharegpt4v: Improving large multi-modal models with better captions.
\newblock {\em arXiv preprint arXiv:2311.12793}, 2023.

\bibitem{young2024yi}
Alex Young, Bei Chen, Chao Li, Chengen Huang, Ge~Zhang, Guanwei Zhang, Heng Li, Jiangcheng Zhu, Jianqun Chen, Jing Chang, et~al.
\newblock Yi: Open foundation models by 01. ai.
\newblock {\em arXiv preprint arXiv:2403.04652}, 2024.

\bibitem{gpt4}
Josh Achiam, Steven Adler, Sandhini Agarwal, Lama Ahmad, Ilge Akkaya, Florencia~Leoni Aleman, Diogo Almeida, Janko Altenschmidt, Sam Altman, Shyamal Anadkat, et~al.
\newblock Gpt-4 technical report.
\newblock {\em arXiv preprint arXiv:2303.08774}, 2023.

\bibitem{resnet}
Kaiming He, Xiangyu Zhang, Shaoqing Ren, and Jian Sun.
\newblock Deep residual learning for image recognition.
\newblock In {\em Proceedings of the IEEE conference on computer vision and pattern recognition}, pages 770--778, 2016.

\bibitem{datacomp}
Samir~Yitzhak Gadre, Gabriel Ilharco, Alex Fang, Jonathan Hayase, Georgios Smyrnis, Thao Nguyen, Ryan Marten, Mitchell Wortsman, Dhruba Ghosh, Jieyu Zhang, et~al.
\newblock Datacomp: In search of the next generation of multimodal datasets.
\newblock {\em Advances in Neural Information Processing Systems}, 36, 2024.

\bibitem{allava}
Guiming~Hardy Chen, Shunian Chen, Ruifei Zhang, Junying Chen, Xiangbo Wu, Zhiyi Zhang, Zhihong Chen, Jianquan Li, Xiang Wan, and Benyou Wang.
\newblock Allava: Harnessing gpt4v-synthesized data for a lite vision-language model.
\newblock {\em arXiv preprint arXiv:2402.11684}, 2024.

\bibitem{objects365}
Shuai Shao, Zeming Li, Tianyuan Zhang, Chao Peng, Gang Yu, Xiangyu Zhang, Jing Li, and Jian Sun.
\newblock Objects365: A large-scale, high-quality dataset for object detection.
\newblock In {\em Proceedings of the IEEE/CVF international conference on computer vision}, pages 8430--8439, 2019.

\bibitem{visualgenome}
Ranjay Krishna, Yuke Zhu, Oliver Groth, Justin Johnson, Kenji Hata, Joshua Kravitz, Stephanie Chen, Yannis Kalantidis, Li-Jia Li, David~A Shamma, et~al.
\newblock Visual genome: Connecting language and vision using crowdsourced dense image annotations.
\newblock {\em International journal of computer vision}, 123:32--73, 2017.

\bibitem{pointqa}
Arjun Mani, Nobline Yoo, Will Hinthorn, and Olga Russakovsky.
\newblock Point and ask: Incorporating pointing into visual question answering.
\newblock {\em arXiv preprint arXiv:2011.13681}, 2020.

\bibitem{flickr30k}
Bryan~A Plummer, Liwei Wang, Chris~M Cervantes, Juan~C Caicedo, Julia Hockenmaier, and Svetlana Lazebnik.
\newblock Flickr30k entities: Collecting region-to-phrase correspondences for richer image-to-sentence models.
\newblock In {\em Proceedings of the IEEE international conference on computer vision}, pages 2641--2649, 2015.

\bibitem{mmc}
Fuxiao Liu, Xiaoyang Wang, Wenlin Yao, Jianshu Chen, Kaiqiang Song, Sangwoo Cho, Yaser Yacoob, and Dong Yu.
\newblock Mmc: Advancing multimodal chart understanding with large-scale instruction tuning.
\newblock {\em arXiv preprint arXiv:2311.10774}, 2023.

\bibitem{chart2text}
Shankar Kantharaj, Rixie~Tiffany Leong, Xiang Lin, Ahmed Masry, Megh Thakkar, Enamul Hoque, and Shafiq Joty.
\newblock Chart-to-text: A large-scale benchmark for chart summarization.
\newblock In {\em Proceedings of the 60th Annual Meeting of the Association for Computational Linguistics (Volume 1: Long Papers)}, pages 4005--4023, 2022.

\bibitem{dvqa}
Kushal Kafle, Brian Price, Scott Cohen, and Christopher Kanan.
\newblock Dvqa: Understanding data visualizations via question answering.
\newblock In {\em Proceedings of the IEEE conference on computer vision and pattern recognition}, pages 5648--5656, 2018.

\bibitem{scigraphqa}
Shengzhi Li and Nima Tajbakhsh.
\newblock Scigraphqa: A large-scale synthetic multi-turn question-answering dataset for scientific graphs.
\newblock {\em arXiv preprint arXiv:2308.03349}, 2023.

\bibitem{llavar}
Yanzhe Zhang, Ruiyi Zhang, Jiuxiang Gu, Yufan Zhou, Nedim Lipka, Diyi Yang, and Tong Sun.
\newblock Llavar: Enhanced visual instruction tuning for text-rich image understanding.
\newblock {\em arXiv preprint arXiv:2306.17107}, 2023.

\bibitem{llava_med}
Chunyuan Li, Cliff Wong, Sheng Zhang, Naoto Usuyama, Haotian Liu, Jianwei Yang, Tristan Naumann, Hoifung Poon, and Jianfeng Gao.
\newblock Llava-med: Training a large language-and-vision assistant for biomedicine in one day.
\newblock {\em Advances in Neural Information Processing Systems}, 36, 2024.

\bibitem{infographicvqa}
Minesh Mathew, Viraj Bagal, Rub{\`e}n Tito, Dimosthenis Karatzas, Ernest Valveny, and CV~Jawahar.
\newblock Infographicvqa.
\newblock In {\em Proceedings of the IEEE/CVF Winter Conference on Applications of Computer Vision}, pages 1697--1706, 2022.

\bibitem{ai2d}
Aniruddha Kembhavi, Mike Salvato, Eric Kolve, Minjoon Seo, Hannaneh Hajishirzi, and Ali Farhadi.
\newblock A diagram is worth a dozen images.
\newblock In {\em Computer Vision--ECCV 2016: 14th European Conference, Amsterdam, The Netherlands, October 11--14, 2016, Proceedings, Part IV 14}, pages 235--251. Springer, 2016.

\bibitem{stvqa}
Ali~Furkan Biten, Ruben Tito, Andres Mafla, Lluis Gomez, Mar{\c{c}}al Rusinol, Ernest Valveny, CV~Jawahar, and Dimosthenis Karatzas.
\newblock Scene text visual question answering.
\newblock In {\em Proceedings of the IEEE/CVF international conference on computer vision}, pages 4291--4301, 2019.

\bibitem{textvqa}
Amanpreet Singh, Vivek Natarajan, Meet Shah, Yu~Jiang, Xinlei Chen, Dhruv Batra, Devi Parikh, and Marcus Rohrbach.
\newblock Towards vqa models that can read.
\newblock In {\em Proceedings of the IEEE/CVF conference on computer vision and pattern recognition}, pages 8317--8326, 2019.

\bibitem{donut}
Geewook Kim, Teakgyu Hong, Moonbin Yim, Jinyoung Park, Jinyeong Yim, Wonseok Hwang, Sangdoo Yun, Dongyoon Han, and Seunghyun Park.
\newblock Donut: Document understanding transformer without ocr.
\newblock {\em arXiv preprint arXiv:2111.15664}, 7:15, 2021.

\bibitem{geo3k}
Pan Lu, Ran Gong, Shibiao Jiang, Liang Qiu, Siyuan Huang, Xiaodan Liang, and Song-Chun Zhu.
\newblock Inter-gps: Interpretable geometry problem solving with formal language and symbolic reasoning.
\newblock {\em arXiv preprint arXiv:2105.04165}, 2021.

\bibitem{pgps9k}
Ming-Liang Zhang, Fei Yin, and Cheng-Lin Liu.
\newblock A multi-modal neural geometric solver with textual clauses parsed from diagram.
\newblock {\em arXiv preprint arXiv:2302.11097}, 2023.

\bibitem{gllava}
Jiahui Gao, Renjie Pi, Jipeng Zhang, Jiacheng Ye, Wanjun Zhong, Yufei Wang, Lanqing Hong, Jianhua Han, Hang Xu, Zhenguo Li, et~al.
\newblock G-llava: Solving geometric problem with multi-modal large language model.
\newblock {\em arXiv preprint arXiv:2312.11370}, 2023.

\bibitem{vqa_rad}
Jason~J Lau, Soumya Gayen, Asma Ben~Abacha, and Dina Demner-Fushman.
\newblock A dataset of clinically generated visual questions and answers about radiology images.
\newblock {\em Scientific data}, 5(1):1--10, 2018.

\bibitem{laion4v}
LAION.
\newblock Gpt-4v dataset.
\newblock \url{https://huggingface.co/datasets/laion/gpt4v-dataset}, 2023.

\bibitem{textocr4v}
Carter Jimmy.
\newblock Textocr-gpt4v.
\newblock \url{https://huggingface.co/datasets/jimmycarter/textocr-gpt4v}, 2024.

\bibitem{gemini}
Gemini Team, Rohan Anil, Sebastian Borgeaud, Yonghui Wu, Jean-Baptiste Alayrac, Jiahui Yu, Radu Soricut, Johan Schalkwyk, Andrew~M Dai, Anja Hauth, et~al.
\newblock Gemini: a family of highly capable multimodal models.
\newblock {\em arXiv preprint arXiv:2312.11805}, 2023.

\bibitem{mplugowl2}
Qinghao Ye, Haiyang Xu, Jiabo Ye, Ming Yan, Haowei Liu, Qi~Qian, Ji~Zhang, Fei Huang, and Jingren Zhou.
\newblock mplug-owl2: Revolutionizing multi-modal large language model with modality collaboration.
\newblock {\em arXiv preprint arXiv:2311.04257}, 2023.

\bibitem{llavanext}
Haotian Liu, Chunyuan Li, Yuheng Li, Bo~Li, Yuanhan Zhang, Sheng Shen, and Yong~Jae Lee.
\newblock Llava-next: Improved reasoning, ocr, and world knowledge, January 2024.

\bibitem{mme}
Chaoyou Fu, Peixian Chen, Yunhang Shen, Yulei Qin, Mengdan Zhang, Xu~Lin, Jinrui Yang, Xiawu Zheng, Ke~Li, Xing Sun, et~al.
\newblock Mme: A comprehensive evaluation benchmark for multimodal large language models.
\newblock {\em arXiv preprint arXiv:2306.13394}, 2023.

\bibitem{qbench}
Haoning Wu, Zicheng Zhang, Erli Zhang, Chaofeng Chen, Liang Liao, Annan Wang, Chunyi Li, Wenxiu Sun, Qiong Yan, Guangtao Zhai, et~al.
\newblock Q-bench: A benchmark for general-purpose foundation models on low-level vision.
\newblock {\em arXiv preprint arXiv:2309.14181}, 2023.

\bibitem{mathvista}
Pan Lu, Hritik Bansal, Tony Xia, Jiacheng Liu, Chunyuan Li, Hannaneh Hajishirzi, Hao Cheng, Kai-Wei Chang, Michel Galley, and Jianfeng Gao.
\newblock Mathvista: Evaluating mathematical reasoning of foundation models in visual contexts.
\newblock {\em arXiv preprint arXiv:2310.02255}, 2023.

\bibitem{mathverse}
Renrui Zhang, Dongzhi Jiang, Yichi Zhang, Haokun Lin, Ziyu Guo, Pengshuo Qiu, Aojun Zhou, Pan Lu, Kai-Wei Chang, Peng Gao, et~al.
\newblock Mathverse: Does your multi-modal llm truly see the diagrams in visual math problems?
\newblock {\em arXiv preprint arXiv:2403.14624}, 2024.

\bibitem{yi}
01-AI.
\newblock Yi.
\newblock \url{https://huggingface.co/01-ai}, 2023.

\bibitem{cogagent}
Wenyi Hong, Weihan Wang, Qingsong Lv, Jiazheng Xu, Wenmeng Yu, Junhui Ji, Yan Wang, Zihan Wang, Yuxiao Dong, Ming Ding, et~al.
\newblock Cogagent: A visual language model for gui agents.
\newblock {\em arXiv preprint arXiv:2312.08914}, 2023.

\bibitem{palix}
Xi~Chen, Josip Djolonga, Piotr Padlewski, Basil Mustafa, Soravit Changpinyo, Jialin Wu, Carlos~Riquelme Ruiz, Sebastian Goodman, Xiao Wang, Yi~Tay, et~al.
\newblock Pali-x: On scaling up a multilingual vision and language model.
\newblock {\em arXiv preprint arXiv:2305.18565}, 2023.

\bibitem{vqav2}
Yash Goyal, Tejas Khot, Douglas Summers-Stay, Dhruv Batra, and Devi Parikh.
\newblock Making the v in vqa matter: Elevating the role of image understanding in visual question answering.
\newblock In {\em Proceedings of the IEEE conference on computer vision and pattern recognition}, pages 6904--6913, 2017.

\bibitem{scienceqa}
Pan Lu, Swaroop Mishra, Tony Xia, Liang Qiu, Kai-Wei Chang, Song-Chun Zhu, Oyvind Tafjord, Peter Clark, and Ashwin Kalyan.
\newblock Learn to explain: Multimodal reasoning via thought chains for science question answering.
\newblock In {\em The 36th Conference on Neural Information Processing Systems (NeurIPS)}, 2022.

\bibitem{ferret}
Haoxuan You, Haotian Zhang, Zhe Gan, Xianzhi Du, Bowen Zhang, Zirui Wang, Liangliang Cao, Shih-Fu Chang, and Yinfei Yang.
\newblock Ferret: Refer and ground anything anywhere at any granularity.
\newblock {\em arXiv preprint arXiv:2310.07704}, 2023.

\bibitem{groundingdino}
Shilong Liu, Zhaoyang Zeng, Tianhe Ren, Feng Li, Hao Zhang, Jie Yang, Chunyuan Li, Jianwei Yang, Hang Su, Jun Zhu, et~al.
\newblock Grounding dino: Marrying dino with grounded pre-training for open-set object detection.
\newblock {\em arXiv preprint arXiv:2303.05499}, 2023.

\bibitem{uninext}
Bin Yan, Yi~Jiang, Jiannan Wu, Dong Wang, Ping Luo, Zehuan Yuan, and Huchuan Lu.
\newblock Universal instance perception as object discovery and retrieval.
\newblock In {\em Proceedings of the IEEE/CVF Conference on Computer Vision and Pattern Recognition}, pages 15325--15336, 2023.

\bibitem{hop}
Zhuofan Zong, Dongzhi Jiang, Guanglu Song, Zeyue Xue, Jingyong Su, Hongsheng Li, and Yu~Liu.
\newblock Temporal enhanced training of multi-view 3d object detector via historical object prediction.
\newblock In {\em Proceedings of the IEEE/CVF International Conference on Computer Vision}, pages 3781--3790, 2023.

\bibitem{dino_det}
Hao Zhang, Feng Li, Shilong Liu, Lei Zhang, Hang Su, Jun Zhu, Lionel~M Ni, and Heung-Yeung Shum.
\newblock Dino: Detr with improved denoising anchor boxes for end-to-end object detection.
\newblock {\em arXiv preprint arXiv:2203.03605}, 2022.

\bibitem{layoutlmv3}
Yupan Huang, Tengchao Lv, Lei Cui, Yutong Lu, and Furu Wei.
\newblock Layoutlmv3: Pre-training for document ai with unified text and image masking.
\newblock In {\em Proceedings of the 30th ACM International Conference on Multimedia}, pages 4083--4091, 2022.

\bibitem{sit}
Zhuofan Zong, Kunchang Li, Guanglu Song, Yali Wang, Yu~Qiao, Biao Leng, and Yu~Liu.
\newblock Self-slimmed vision transformer.
\newblock In {\em European Conference on Computer Vision}, pages 432--448. Springer, 2022.

\bibitem{lisa}
Xin Lai, Zhuotao Tian, Yukang Chen, Yanwei Li, Yuhui Yuan, Shu Liu, and Jiaya Jia.
\newblock Lisa: Reasoning segmentation via large language model.
\newblock {\em arXiv preprint arXiv:2308.00692}, 2023.

\bibitem{visionllm}
Wenhai Wang, Zhe Chen, Xiaokang Chen, Jiannan Wu, Xizhou Zhu, Gang Zeng, Ping Luo, Tong Lu, Jie Zhou, Yu~Qiao, et~al.
\newblock Visionllm: Large language model is also an open-ended decoder for vision-centric tasks.
\newblock {\em Advances in Neural Information Processing Systems}, 36, 2024.

\end{thebibliography}

\newpage
\appendix

\section{Appendix / supplemental material}

\begin{table}[h]
    \centering
    \caption{\textbf{Vision expert model configurations of vision experts in MoVA.} Methods with * use a convolution layer to compress the output feature.}
    \label{tab:vision_expert}
    \setlength\tabcolsep{2pt}
    \resizebox{0.75\textwidth}{!}{
        \begin{tabular}{@{}l|ccccc@{}}
        \toprule
        Model & Params & Resolution & Width & Depth & Output shape \\
        \midrule
        DINOv2-giant~\cite{dinov2} & 1.1B & 518$\times$518 & 1536 & 40 & 1536$\times$37$\times$37 \\
        Co-DETR-large*~\cite{codetr} & 304M & 1280$\times$1280 & 1024 & 24 & 256$\times$80$\times$80 \\        
        SAM-huge*~\cite{sam} & 632M & 1024$\times$1024 & 1280 & 32 & 256$\times$64$\times$64 \\
        Pix2Struct-large~\cite{pix2struct} & 513M & 720$\times$720 & 1536 & 18 & 1536$\times$45$\times$45 \\
        Deplot-base~\cite{deplot} & 92M & 720$\times$720 & 768 & 12 & 768$\times$45$\times$45 \\
        Vary-base*~\cite{dinov2} & 86M & 1024$\times$1024 & 768 & 12 & 512$\times$32$\times$32  \\
        BiomedCLIP-base~\cite{dinov2} & 86M & 224$\times$224 & 768 & 12 & 768$\times$16$\times$16 \\
        \bottomrule
        \end{tabular}       
    }
\end{table}

\begin{table}[h]
    \centering
    \caption{\textbf{Introduction of datasets used in the MoV-Adapter pretraining stage.} The <class> placeholder represents the object category in the object detection task. The <expr> placeholder represents the expression in the REC task. The <bbox> placeholder denotes the bounding box coordinates. The <point> placeholder denotes the coordinate of a point. We directly use the original question as the instruction for MMC-Instruction and ScigraphQA.}
    \label{tab:training_data}
    \setlength\tabcolsep{3pt}
    \resizebox{\textwidth}{!}{
        \begin{tabular}{@{}l|ll@{}}
        \toprule
        Task & Dataset & Task template \\
        \midrule
        \multirow{6}{*}{Image Caption} & \multirow{2}{*}{Datacomp~\cite{datacomp}} & Please describe this image. \\
         & & Provide a one-sentence caption for the provided image. \\
        \cmidrule(lr){2-3}
         & \multirow{2}{*}{ShareGPT4V-PT~\cite{sharegpt4v}} & Can you elaborate on the elements of the picture provided? \\
         & & Write a detailed description of the given image. \\
         \cmidrule(lr){2-3}
         & \multirow{2}{*}{ALLaVA-4V~\cite{allava}} & Can you elaborate on the elements of the picture provided? \\
         & & Write a detailed description of the given image. \\         
         \cmidrule(lr){1-3}
        \multirow{10}{*}{Grounding and Localization} & \multirow{2}{*}{Objects365~\cite{objects365}} & Detect all objects among <class> in the image. \\         & & Perform object detection given the image within <class>. \\
        \cmidrule(lr){2-3}
         & \multirow{2}{*}{RefCOCO~\cite{refcoco}} & Locate the region this sentence describes: <expr>. Please provide the bounding box coordinates. \\         & & Please generate a short and spotlighted mention of the <bbox> part seen in the photo. \\    
        \cmidrule(lr){2-3}
         & \multirow{2}{*}{Visual Genome~\cite{visualgenome}} & Locate the region this sentence describes: <expr>. Please provide the bounding box coordinates. \\         & & Please generate a short and spotlighted mention of the <bbox> part seen in the photo. \\        
        \cmidrule(lr){2-3}
         & \multirow{2}{*}{PointQA~\cite{pointqa}} & How many of these objects <bbox> in picture? \\
         & & How many of these objects <point> in picture? \\    
        \cmidrule(lr){2-3}
         & \multirow{2}{*}{Flickr30K~\cite{flickr30k}} & Take a look at the image and give me the location details for any mentioned items. \\
         & & Unravel the aspects of the image and give the bounding box for the mentioned items. \\    
         \cmidrule(lr){1-3}
        \multirow{4}{*}{Chart Understanding} & MMC-Instruction~\cite{mmc} & - \\
        \cmidrule(lr){2-3}
         & \multirow{2}{*}{Chart2Text~\cite{chart2text}} & What significant details and conclusions can be drawn from this chart? \\
        & & Can you extract the data points in this image? \\  
        \cmidrule(lr){2-3}
        & ScigraphQA~\cite{scigraphqa} & - \\
        \cmidrule(lr){1-3}
        \multirow{4}{*}{Document Parsing} & \multirow{2}{*}{LLaVAR-PT~\cite{llavar}} & Report on any text that can be clearly read in the image. \\  
        & & Identify any text visible in the image provided. \\  
        \cmidrule(lr){2-3}
         & \multirow{2}{*}{English Documents} & Extract every piece of text from this image. \\
         & & I request you to apply optical character recognition to this image. \\  
         \cmidrule(lr){1-3}   
        \multirow{2}{*}{Biomedical Understanding} & \multirow{2}{*}{LLaVA-Med~\cite{llava_med}} & Write a terse but informative summary of the picture. \\       
         & & Share a comprehensive rundown of the presented image. \\ 
        \bottomrule
        \end{tabular}       
    }
\end{table}

\begin{table*}[t]
    \begin{minipage}[t]{0.33\linewidth}
        \centering
        \caption{Performance of various routing component.
        }
        \resizebox{\textwidth}{!}{
            \begin{tabular}{@{}c|c|ccc@{}}
            \toprule
            Design & \#Samples & MME & MMB \\
            \midrule
            \cellcolor{gray!15}\textbf{LLM} & \cellcolor{gray!15}{2K} & \cellcolor{gray!15}\textbf{1562/371} & \cellcolor{gray!15}\textbf{70.4} \\
            \midrule
            BERT & \textbf{1.6M} & 1520/326 & 68.8 \\
            MLP & \textbf{1.6M} & 1483/305 & 68.1 \\
            \bottomrule
            \end{tabular}
        }
        \label{tab:ablation_llm}
    \end{minipage}
    \hspace{0.2cm}
    \begin{minipage}[t]{0.39\linewidth}
        \centering
        \caption{
        Results of various vision encoder combination for routing.
        }
        \resizebox{\textwidth}{!}{
            \begin{tabular}{@{}c|ccc@{}}
            \toprule
            Design & MMB & GQA & DocVQA \\
            \midrule
            \cellcolor{gray!15}\textbf{CLIP} & \cellcolor{gray!15}\textbf{70.4} & \cellcolor{gray!15}{64.8} & \cellcolor{gray!15}\textbf{81.3} \\
            \midrule
            +DINOv2 & 70.1  & \textbf{65.1} & 80.5 \\
            +DINOv2+Pix2Struct & 69.5 & 64.4 & 80.9 \\
            \bottomrule
            \end{tabular}
        }
        \label{tab:ablation_routing_vit}
    \end{minipage}
    \hspace{0.2cm}
    \begin{minipage}[t]{0.22\linewidth}
        \centering
        \caption{Effects of routing image tokens. 
        }
        \resizebox{\textwidth}{!}{
            \begin{tabular}{@{}c|ccccc@{}}
            \toprule
            \#IMG & MMB & ChartQA \\
            \midrule
            \cellcolor{gray!15}144 & \cellcolor{gray!15}{70.4} & \cellcolor{gray!15}{68.3} \\
            \midrule
            256 & 69.8 & 68.4 \\
            \textbf{576} & \textbf{70.6} & \textbf{68.7} \\
            \bottomrule
            \end{tabular}
        }
        \label{tab:ablation_routing_token}
    \end{minipage}    
    \vspace{-0.3cm}
\end{table*}
\begin{table*}[t!]
\centering
\caption{\textbf{Model descriptions used in the context-aware expert routing.} We describe the pros and cons of each expert model in the routing prompt. We only present 3 captions for each expert here.}
\setlength\tabcolsep{3pt}
\begin{minipage}{1.0\columnwidth}
\vspace{0mm}    
\centering
\begin{tcolorbox} 
    \centering
      \footnotesize
    \begin{tabular}{p{0.97\columnwidth} c}
   \VarSty{ {\bf {DINOv2 description}} } &\\
(1) This model demonstrates exceptional prediction capabilities across a range of image-related tasks, including image classification, object detection, segmentation, and image retrieval. The model leverages advanced self-supervised learning techniques to achieve high performance without relying heavily on labeled data.&  \\
(2) This model shows very strong prediction capabilities on tasks such as image classification, detection, segmentation, and image retrieval. However, it encounters challenges in accurately reading text within images.&  \\
(3) This model can effectively extract the accurate spatial and semantic information from natural images.&  \\
    \hrulefill & \\
   \VarSty{ {\bf {Co-DETR description}} } &\\
(1) This model is a state-of-the-art object detector pretrained on natural images. It can enable models to solve object-centric problems. Nonetheless, this model struggles with processing background elements in natural scenes. & \\ 
(2) This model is a cutting-edge object detection model that can accurately detect objects in images. However, it struggles with identifying text in images. & \\
(3) This model is a state-of-the-art object detector that can identify objects in images. & \\
    \hrulefill & \\
   \VarSty{ {\bf {SAM description}} } &\\
(1) This model is an image segmentation model. This model can segment the precise location of either specific objects in an image or every object in an image.&  \\  
(2) This model is a leading image segmentation framework and achieves strong zero-shot segmentation performance. & \\
(3) This model is a promotable segmentation system with zero-shot generalization to unfamiliar objects and images. & \\
    \hrulefill & \\
   \VarSty{ {\bf {Pix2Struct description}} } &\\
(1) This model excels in text recognition, achieving state-of-the-art text analysis results across distinct domains: documents, illustrations, user interfaces, natural images containing text, and images of charts. &  \\  
(2) This model demonstrates exceptional proficiency in text recognition, delivering cutting-edge text analysis performance across various domains.& \\
(3) This model can automate the extraction of information from scanned documents, making it easier to digitize and manage large volumes of paperwork. &\\
    \hrulefill & \\
   \VarSty{ {\bf {Deplot description}} } &\\
(1) This model is a specialized model designed to achieve state-of-the-art plot and chart understanding performance.&  \\  
(2) This model is a fine-tuned version of an existing text recognition model. It has been specifically trained to achieve superior performance in plot and chart understanding tasks.& \\
(3) This model can help detect the text within the input document, diagram, and chart images. &\\
    \hrulefill & \\
   \VarSty{ {\bf {Vary description}} } &\\
(1) This model can achieve more fine-grained vision perception for images with text, such as document-level Chinese/English OCR, book image to markdown or LATEX, Chinese/English chart understanding.&  \\  
(2) This model can handle images with text effectively and accurately, enabling advanced tasks such as document OCR and chart understanding.& \\
(3) This model can accurately process images with text, enabling tasks such as OCR. However, it cannot process natural images without text. & \\
    \hrulefill & \\
   \VarSty{ {\bf {BiomedCLIP description}} } &\\
(1) This model is a foundation model designed for biomedical vision-language processing.&  \\  
(2) This model is capable of biomedical images, such as chest X-ray and radiology images.& \\
(3) This model is a state-of-the-art biomedical vision-language model. It has been shown to achieve significant improvements in biomedical image-text tasks.& \\
    \end{tabular}
\end{tcolorbox}
    \label{tab:descriptions}    
\end{minipage}
\end{table*}

\subsection{Vision Experts}
\label{sec:appendix_vision}
\textbf{Model Configuration.}
We present the detailed model configurations of our task-specific vision experts in~\ref{tab:vision_expert}.
We adopt the official checkpoint weights that are publicly available.

\textbf{Model Description.}
The model descriptions used in the routing prompt are released in Table~\ref{tab:descriptions}.

\subsection{Training Data Details}
The training process of MoVA consists of two stages.
In the Appendix, we present the training datasets with corresponding task templates of the first stage in Table~\ref{tab:training_data}.
For the training data of the second stage, we follow the prompt format of LLaVA-1.5~\cite{llava15}.
The MoVA models with Vicuna-7B and LLama3-8B are pretrained using 64 A100 80G GPUs for 2 days, and finetuned using 32 A100 80G GPUs for 1 day.
The MoVA with 34B LLM is pretrained using 128 A100 80G GPUs for 5 days and finetuned using 64 A100 80G GPUs for 2 days.

\subsection{More Experiments}
\label{sec:appendix_exp}

\textbf{Image segmentation.}
In this experiment, we aim to investigate if task-specific knowledge can improve MoVA on the segmentation task.
Therefore, we introduce a simple design to extend MoVA to segmentation tasks.
Unlike segmentation generalists~\cite{lisa} that adopt an additional pixel decoder with high-resolution images for high-quality mask generation, we just formulate the referring segmentation task as sequential polygon generation~\cite{visionllm}.
We finetune MoVA and the baseline with a SAM-Huge~\cite{sam} backbone on the RefCOCO referring segmentation datasets.
MoVA achieves 57.1\% gIoU on the testA benchmark, which is 2.6\% higher than the 54.5\% of baseline.
This result indicates that MoVA is capable of exploiting task-specific knowledge to solve segmentation tasks.


\textbf{Effect of LLM for expert routing.}
In this experiment, we investigate the effect of the LLM in our coarse-grained expert routing.
As presented in Table~\ref{tab:ablation_llm}, expert routing with LLM achieves the best performance.
When the LLM is substituted for a lightweight MLP classifier and a BERT encoder, we need to increase the number of routing training samples from 2K to 1.6M to preserve model performance.
Besides, both MLP classifier and BERT encoder fail to perform expert routing in open scenarios as stated in Table~\ref{tab:ablation_open}.
Therefore, the strong tool-use capacity and generalization ability of LLM is critical to our flexible and effective expert routing.

\textbf{Vision encoder for expert routing.}
In the coarse-grained expert routing, we only adopt the image feature of base vision encoder CLIP.
As presented in Table~\ref{tab:ablation_routing_vit}, the method with CLIP achieves slightly better performance than other methods since such a routing task does not require elaborate expert knowledge.
Besides, incorporating other vision experts with plain fusion also brings biased information and increases cost.
To achieve a better trade-off between efficiency and performance, we only use CLIP for coarse-grained expert routing.

\textbf{Number of image tokens in expert routing.}
We ablate the number of image tokens used in the expert routing stage.
As shown in Table~\ref{tab:ablation_routing_token}, routing with 144 tokens can achieve comparable performance to methods with more tokens.
Considering the additional cost brought by processing more image tokens, we only use 144 tokens for routing.

\textbf{Data construction.}
In this experiment, we analyze the effectiveness of our routing data construction method.
The test split is constructed by randomly selecting 500 samples from the routing data.
We first prompt GPT4-V~\cite{gpt4} to determine whether the expert routing result is reasonable or not.
These evaluation results are then manually checked and we finally compute the routing accuracy using the human-verified evaluation results.
Our loss-driven data construction approach achieves 94.6\% accuracy (473 samples are regarded as ``reasonable'' by GPT4-V), demonstrating its effectiveness.

\subsection{Qualitative Results}
\begin{figure*}[tp]
    \centering
    \includegraphics[width=\textwidth]{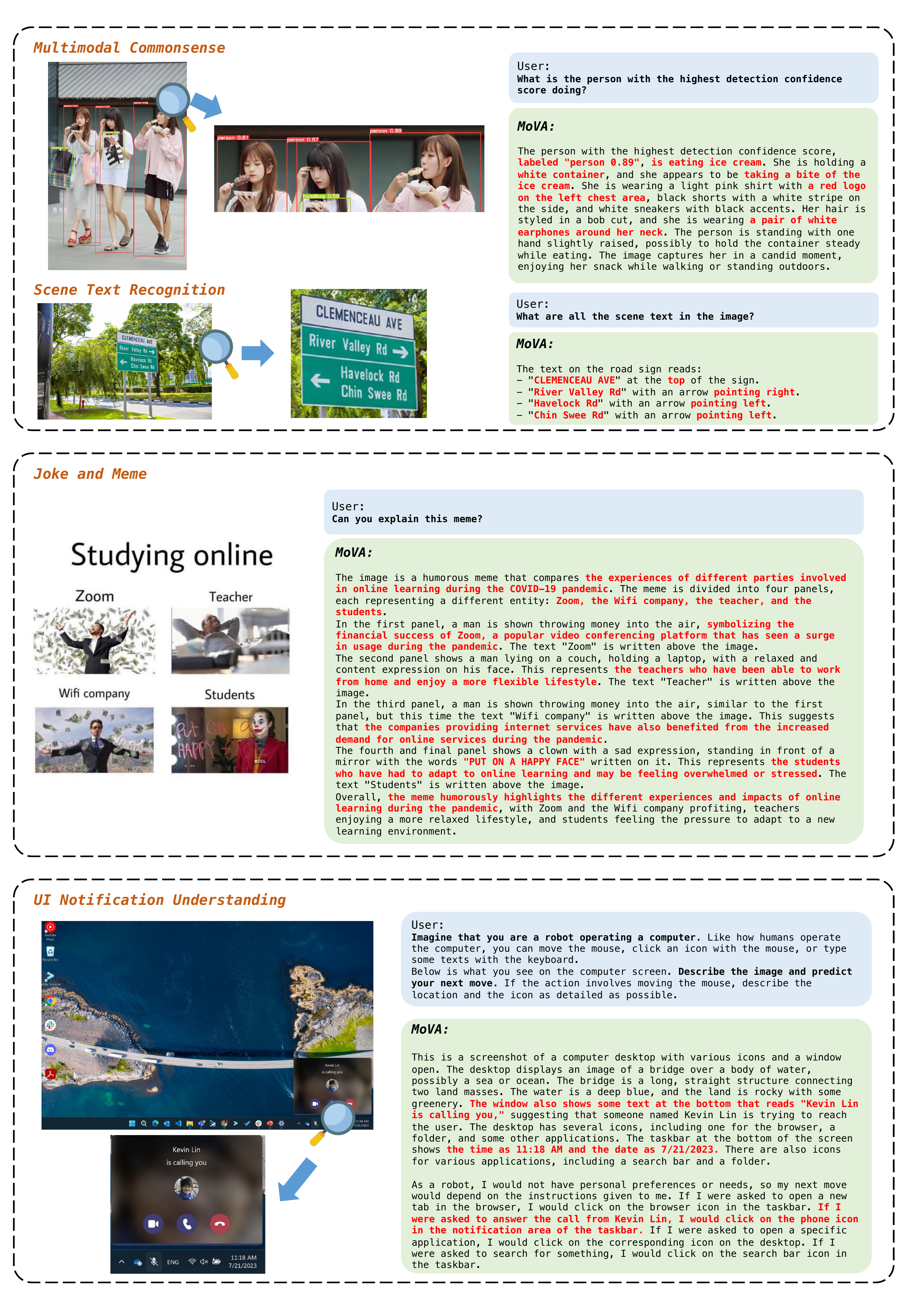}
    \caption{Qualitative multimodal understanding results of MoVA.
    }
    \label{fig:results1}
\end{figure*}

\begin{figure*}[tp]
    \centering
    \includegraphics[width=\textwidth]{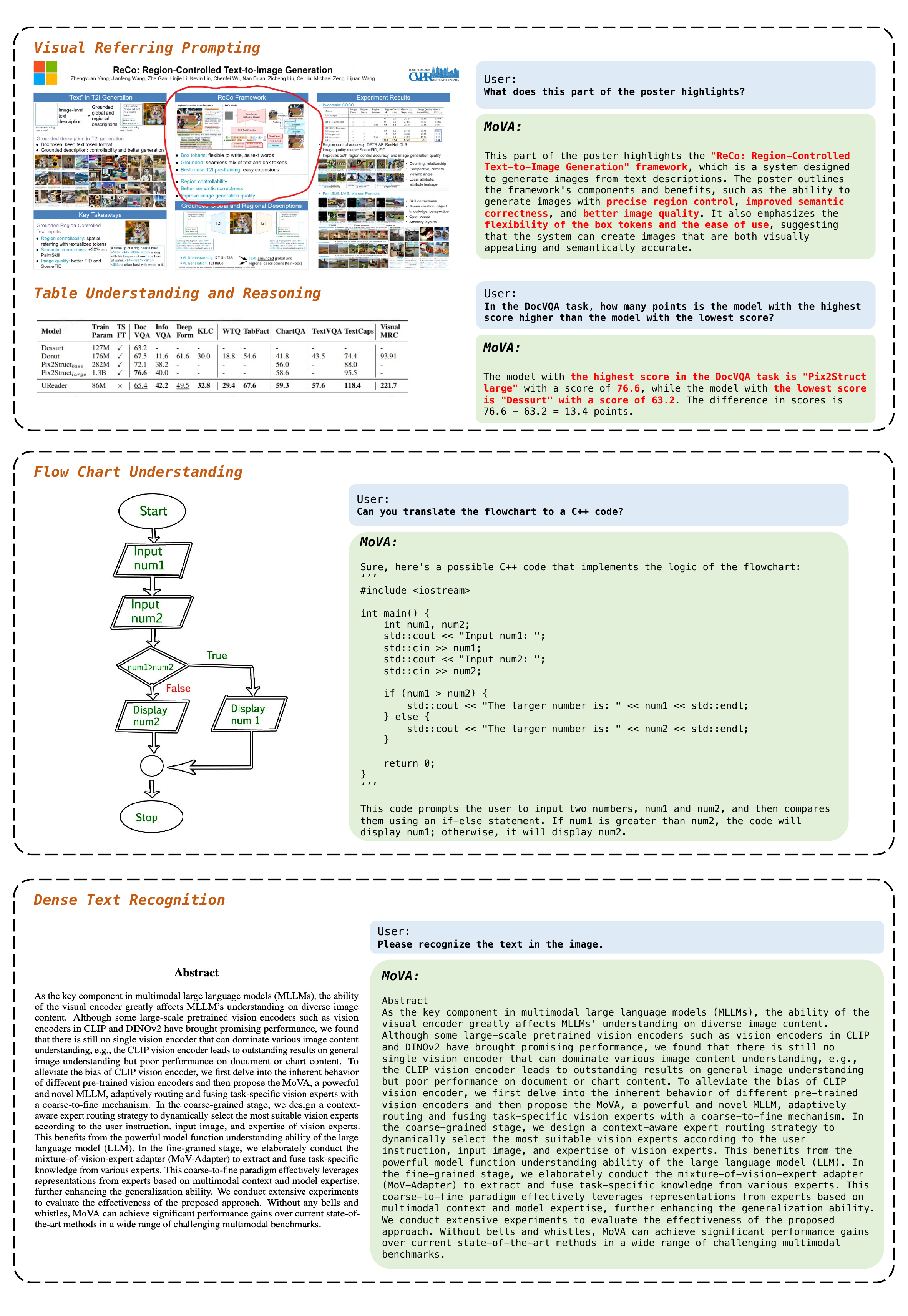}
    \caption{Qualitative multimodal understanding results of MoVA.
    }
    \label{fig:results2}
\end{figure*}

We apply MoVA to a wide range of real-world understanding and reasoning tasks to investigate its multimodal comprehension capacity.
As presented in Figure~\ref{fig:results1} and Figure~\ref{fig:results2}, MoVA successfully solves these diverse and complex cases thanks to the context-relevant expert knowledge.

\subsection{Potential Societal Impacts}
\textbf{Positive impacts.}
The positive impact is to use MoVA to solve various open-world tasks in real-world scenarios.
For example, MoVA can serve as a useful assistant to help people improve work efficiency and answer their questions.

\textbf{Negative impacts.}
The potential negative social impact is to use MoVA model to generate misleading and false contents.
This issue potentially presents in all multimodal large language models.
We will try to improve its safety in future works.


\end{document}